\documentclass[twoside]{article}

\usepackage{hyperref} 
\usepackage{amsmath}
\usepackage{amssymb}
\usepackage{mathtools}
\usepackage{amsthm}
\usepackage{comment}
\usepackage{multirow}
\usepackage{booktabs}
\usepackage{subcaption}
\usepackage{array}
\usepackage{algorithm}
\usepackage{algpseudocode}
\usepackage{xcolor}
\theoremstyle{plain}

\theoremstyle{definition}

\theoremstyle{remark}

\usepackage[preprint]{aistats2027}
\usepackage[round]{natbib}

\begin{document}

%

%

\twocolumn[

\aistatstitle{Spectral Regularization for Diffusion Models}

\aistatsauthor{Satish Chandran$^{*1}$ \And N\'icolas Roque dos Santos$^{*2}$ \And Yunshu Wu$^{2}$}
\aistatsauthor{Greg Ver Steeg$^{2}$ \And Evangelos Papalexakis$^{2}$}

\aistatsaddress{$^{1}$Department of Mathematics, University of California, Riverside, CA, USA \\
  $^{2}$Department of Computer Science, University of California, Riverside, CA, USA \\[4pt]
  \texttt{\{schan360, nicolasr, ywu380, gregoryv, epapalex\}@ucr.edu}}
]

{\renewcommand{\thefootnote}{\fnsymbol{footnote}}%
 \footnotetext[1]{Equal contribution.}}

\begin{abstract}
Diffusion models are commonly trained with pointwise objectives that do not explicitly distinguish spectral or multiscale structure. We introduce a spectral regularization framework that augments training with discrepancies in Fourier amplitudes and/or phase, wavelet coefficients, or graph Laplacian eigenvalues. The losses act on clean predictions or final distillation states and leave the architecture, diffusion process, and sampler unchanged. We characterize how reconstruction parameterization scales spectral-loss sensitivity and evaluate the framework in fine-tuning and distillation. In paired experiments on CIFAR-10 and FFHQ, Fourier amplitude regularization modestly improves the FID of students using two network evaluations, whereas both tested wavelet losses degrade performance at every evaluated weight. Fine-tuning improves audio metrics relative to a released DiffWave checkpoint, while molecular graph results exhibit trade-offs across metrics and sampling budgets.
\end{abstract}

\section{Introduction}
Diffusion models have emerged as a powerful and versatile framework for generative modeling of high-dimensional signals. By learning to reverse a gradual noising process, diffusion models provide stable training, strong mode coverage, and state-of-the-art performance across a wide range of modalities, including natural images, audio waveforms, and graphical representations \citep{sohldickstein2015deepunsupervisedlearningusing, ho2020denoisingdiffusionprobabilisticmodels, dhariwal2021diffusionmodelsbeatgans, liu2023generativediffusionmodelsgraphs}. Their success has led to rapid adoption in image synthesis \citep{rombach2022highresolutionimagesynthesislatent}, audio generation \citep{kong2021diffwaveversatilediffusionmodel}, and conditional generation tasks such as super-resolution and inpainting \citep{saharia2022photorealistictexttoimagediffusionmodels}.

Despite their strong empirical performance, diffusion models are typically trained using pointwise reconstruction losses defined in the signal domain, most commonly mean-squared error on predicted noise or clean signals \citep{ho2020denoisingdiffusionprobabilisticmodels,song2021scorebasedgenerativemodelingstochastic}. While these objectives are well motivated from both an empirical and theoretical perspective, they are agnostic to the spectral and multi-scale structure that characterizes many natural signals. Images and audio often exhibit highly structured frequency content, long-range correlations, and scale-dependent patterns that are only implicitly captured through pixel/sample-level supervision. As a result, diffusion models often generate samples that match low-level statistics while exhibiting artifacts such as over-smoothing, incorrect frequency balance, or degraded fine-scale structure \citep{benita2025spectralanalysisdiffusionmodels, chen2025physiowavemultiscalewavelettransformerphysiological, ding2023patcheddenoisingdiffusionmodels, jiralerspong2025shapinginductivebiasdiffusion}.

Recent works have explored augmenting diffusion models with additional forms of structure or inductive bias. In scientific and engineering domains, this has included incorporating explicit physical constraint or residual losses into diffusion training to enforce known properties of the data-generating process \citep{Shu_2023, bastek2025physicsinformeddiffusionmodels, jacobsen2024cocogenphysicallyconsistentconditionedscorebased}. While effective in settings where such constraints are well defined, these approaches are less directly applicable to general image and audio tasks, where the structure is implicit, statistical, and perceptual rather than being defined explicitly as a set of differential equations.

In contrast, frequency-domain representations have long played a central role in image and audio processing. Fourier analysis provides a global description of signal energy distribution across frequencies and is fundamental to understanding smoothness, noise, and periodic structure \citep{Oppenheim_SignalsAndSystems}. Wavelet representations offer a localized, multi-resolution decomposition that capture both spatial/temporal locality and scale \citep{Mallat_Wavelet}. These representations underpin classical methods in compression, denoising, and texture analysis, and they have also been used as inductive biases or auxiliary losses in deep learning models for images and audio \citep{bruna2012invariantscatteringconvolutionnetworks, Gatys_ImageStyleTransfer, yamamoto2020parallelwavegan}.

We propose a spectral regularization framework for diffusion model training that augments the standard denoising objective with differentiable losses defined in the Fourier and wavelet domains. The losses act on the model's prediction of the clean sample, so they leave the forward process, the architecture, and the sampler unchanged, require no auxiliary networks or additional supervision, and only cost one FFT or wavelet transform per training step. Fourier-based losses capture global spectral characteristics, while wavelet-based losses provide localized, scale-aware control. By Parseval’s identity, squared error on full complex Fourier coefficients is equivalent to squared error in the signal domain. Amplitude matching changes this discrepancy by removing phase sensitivity. We use $L^1$ penalties to reduce the relative dominance of large coefficient errors compared with squared penalties.

Our contributions are threefold.
\begin{itemize}
\item We introduce a common framework for augmenting diffusion objectives with Fourier, wavelet, and graph eigenvalue discrepancies. The framework preserves the architecture and sampling procedure. We derive sensitivity bounds that characterize how errors in the network output propagate to spectral losses under noise prediction and EDM preconditioning. 
\item We evaluate Fourier and wavelet regularization through paired few-step distillation experiments on CIFAR-10 and FFHQ. Fourier amplitude regularization improves FID at suitable weights, including a weight transferred from CIFAR-10 to FFHQ without further selection. Both tested wavelet losses degrade performance across the evaluated weights. We further show that improved agreement with the teacher’s average power spectrum does not necessarily improve FID.
\item We examine spectral regularization through fine-tuning experiments on audio, image, and molecular graph models. Comparisons with released checkpoints show improvements in conditional speech synthesis and selected image settings, while graph experiments reveal trade-offs across evaluation metrics and sampling budgets.
\end{itemize}

\section{Related Works} \label{sec:related_work}
\providecommand{\revised}[1]{\ifdefined\textcolor\textcolor{red}{#1}\else#1\fi}

\paragraph{Spectral-domain diffusion models.}
A growing body of work explores incorporating frequency structure directly into diffusion models by redefining the diffusion process in spectral coordinates or operating on transformed representations.~\citet{crabbe2024timeseriesdiffusionfrequency} formulate diffusion for time series in the Fourier domain, while preserving conjugate symmetry of complex coefficients.~\citet{jiralerspong2025shapinginductivebiasdiffusion} shape the diffusion dynamics in frequency space to emphasize or suppress specific spectral bands.~\citet{phillips2022spectraldiffusionprocesses} propose spectral process diffusion, performing score-based modeling over coefficients of stochastic processes expressed in a spectral basis. \citet{falck2025fourier} observe that images, audio, and similar modalities have a Fourier variance that decays by orders of magnitude with frequency. Under additive white noise the standard forward process lowers the signal-to-noise ratio of high frequencies faster than that of low ones, which they show degrades high-frequency generation, and they propose a Fourier-space forward process that corrupts all frequencies at the same rate. \citet{benita2025spectralanalysisdiffusionmodels} model the data as Gaussian with a circulant covariance, which turns the reverse process into a closed-form spectral transfer function, and design noise schedules matched to the data spectrum. \citet{esteves2026spectrally} derive per-image noise schedules from
the power spectrum of each image, with bounds on the useful minimum and maximum
noise levels, which improves pixel diffusion in the few-step regime. These methods change the forward process, its schedule, or the coordinates in which it runs. Our losses leave all three unchanged and act on the training objective instead.
 
\paragraph{Frequency-domain losses.}
Losses on Fourier coefficients have been applied to other machine learning models.
\citet{durall2020watch} add a spectral regularizer on the azimuthally integrated power spectrum of GAN outputs. The focal frequency loss \citep{jiang2021focal} focuses reconstruction and synthesis on frequencies that are hard to fit.  \citet{fuoli2021fourier} combine Fourier-space supervision with a Fourier-domain
discriminator for super-resolution. \citet{yan2024fourier} replace the MSE of precipitation now-casting models with a Fourier amplitude loss and a spectral correlation loss. Applied to the noise prediction of a diffusion model, their loss gave no gain, as the target there is white Gaussian noise. We instead
apply spectral terms to the clean prediction, where the target
has a meaningful approximation of the spectrum of the data. Several related objectives discard phase or position by construction. The multi-resolution STFT loss, introduced as an auxiliary objective for a GAN vocoder \citep{yamamoto2020parallelwavegan}, compares only spectrogram magnitudes and so ignores phase within each analysis window. The Gram matrices of \citet{Gatys_ImageStyleTransfer} are spatially averaged second-order feature statistics, and scattering coefficients \citep{bruna2012invariantscatteringconvolutionnetworks} average wavelet moduli over position to obtain invariance to translation and stability to small deformations. Our amplitude and eigenvalue losses are invariant summaries of the same kind.

\paragraph{Wavelet-based and multi-resolution diffusion.}
Wavelet representations have motivated several diffusion models that operate directly on multi-scale decompositions.~\citet{guth2022waveletscorebasedgenerativemodeling} perform score-based diffusion on wavelet coefficients and interpret the resulting hierarchy through renormalization group theory.~\citet{phung2023waveletdiffusionmodelsfast} and~\citet{hui2022neuralwaveletdomaindiffusion3d} apply wavelet-domain diffusion for efficient image generation and 3D shape modeling, respectively. Related approaches selectively apply diffusion to low-frequency components while refining high-frequency content using auxiliary modules, primarily for restoration and efficiency \citep{huang2024wavedmwaveletbaseddiffusionmodels,zhao2023waveletbasedfourierinformationinteraction,LIU2025104662,ZHOU2026113060}. These methods generally redefine the representation space or introduce task-specific architectural decompositions. Closer to our setting are methods that leave the architecture and sampler untouched and use wavelets only in the training objective. Diffusion-4K \citep{zhang2025diffusion4k} adds a wavelet loss on latent sub-bands when fine-tuning latent diffusion models at 4K resolution. Latent Wavelet Diffusion \citep{sigillo2026latent} instead keeps the flow-matching residual in its original coordinates and uses the high-frequency Haar energy of the latent as a spatial saliency map, which defines a time-dependent binary mask so that detail-rich regions are supervised over more time-steps. In both, wavelets act on autoencoder latents for high-resolution text-to-image synthesis. Our wavelet terms differ in kind, since we penalize the discrepancy between image wavelet coefficients directly in sample space. 

\paragraph{Hybrid Fourier-wavelet diffusion.}
Several recent works explicitly combine Fourier and wavelet representations within diffusion pipelines.~\citet{luo2025waveletfourierdiffuserfrequencyaware} introduce cross-frequency fusion for trajectory modeling in reinforcement learning, while~\citet{kiruluta2025hybridwaveletfouriermethodnextgeneration} propose a hybrid forward process combining partial Fourier corruption with wavelet decomposition and multi-branch denoising networks. Unlike these approaches, which integrate spectral structure into the forward or sampling process, we retain the standard diffusion dynamics and instead regularize training via spectral losses.

\paragraph{Few-step distillation.}
Distillation compresses a many-step sampler into a student that needs only a few network evaluations. The student can regress onto teacher outputs or trajectories \citep{luhman2021knowledge,salimans2022progressive,zhou2024sfd}, enforce self-consistency along the probability flow \citep{song2023consistency}, or match the output distribution through score-based or adversarial objectives \citep{yin2024one,sauer2024adversarial}. In the regression family the distance between student and teacher is a design choice with a large effect. Consistency models first used LPIPS \citep{song2023consistency} and later a Pseudo-Huber distance \citep{song2024improved}, and SFD \citep{zhou2024sfd} compares squared $L^2$, $L^1$, LPIPS, and Pseudo-Huber distances and adopts $L^1$. All of these are pixel-space or learned perceptual distances. Frequency-aware terms have appeared in neighboring settings. Wavelet knowledge distillation \citep{zhang2022wavelet} transfers only the high-frequency bands when compressing GANs, HiPA \citep{zhang2023hipa} adds a perceptual loss on high-pass and edge-filtered images when adapting text-to-image models to one step, and \citet{wang2026spectral} match the distributions of Fourier amplitude and phase of features with a sliced Wasserstein distance in one-step super-resolution. We add the same transform-domain $L^1$ terms used for denoising to the final state of an SFD student, leave its trajectory loss and sampler unchanged, and compare Fourier and wavelet terms under paired seeds. The representation matters, with a translation-invariant amplitude term improving 2-NFE students while localized wavelet terms degrade them.

\paragraph{Spectral structure in graph diffusion.}
Diffusion models for graphs either denoise adjacency matrices under continuous noise \citep{niu2020permutation,jo2022score,yan2024swingnn} or edit discrete node and edge types \citep{vignac2023digress}. Several works bring the graph spectrum into this pipeline. The GAN-based SPECTRE \citep{martinkus2022spectre} first generates the dominant Laplacian eigenvalues and eigenvectors and then builds a graph that matches them, and DiGress supplies spectral features to the denoiser as auxiliary inputs. GSDM \citep{luo2024fast} confines the noise to the eigenvalues of the adjacency matrix, \citet{minello2025generating} and \citet{huang2025graph} generate a truncated set of Laplacian eigenvalues and eigenvectors by diffusion and by flow matching, and Wave-GD \citep{cho2023multiresolution} runs score-based diffusion in a graph wavelet space. These methods change the state space of the generative process or the inputs of the network. At the level of the objective, \citet{siami2025generative} reweights the per-edge diffusion loss by effective resistance, which changes how much each edge counts but not what is matched. We instead keep the adjacency-space EDM formulation of SwinGNN \citep{yan2024swingnn} and penalize the discrepancy between the sorted Laplacian eigenvalues of the clean prediction and those of the data. Laplacian spectra are already a standard evaluation statistic for graph generators \citep{liao2019efficient}, and here they enter the training objective.

\paragraph{Constraint-augmented diffusion models.}
More broadly, diffusion models have been augmented with auxiliary losses to encode known structure, particularly in scientific domains. Physics-informed diffusion models incorporate residual-based constraints derived from governing equations to enforce physical consistency \citep{Shu_2023,bastek2025physicsinformeddiffusionmodels,jacobsen2024cocogenphysicallyconsistentconditionedscorebased}. While effective when explicit constraints are available, these methods are less applicable to natural images and audio, where structure is statistical and perceptual rather than rule-based. Our work draws inspiration from this paradigm but replaces explicit constraints with soft, frequency-domain regularization.

\paragraph{Positioning of Our Approach.} Overall, existing frequency-aware diffusion methods typically modify the diffusion process, operate in transformed domains, or introduce task-specific architectures. In contrast, we propose a loss-level spectral regularization framework that preserves standard diffusion formulations (DDPM, DDIM, and EDM) while encouraging frequency balance and multi-scale coherence through Fourier, wavelet, and graph-spectral losses. This design is domain-agnostic, architecture-independent, and directly compatible with existing diffusion training and sampling pipelines. The amplitude and eigenvalue losses leave the placement of structure to the pointwise objective. The wavelet losses do not, and for an orthonormal wavelet such as Haar, with uniform weights $\gamma_{s,\ell}$ and a squared norm, they would reduce to the pointwise squared error by Parseval's identity. They therefore act as a scale-weighted, localized variant of a constraint the objective already imposes.

\section{Diffusion Models} \label{sec:diffusion_models}
Diffusion models generate data by learning to reverse a gradual noising process. We build on three closely related formulations, denoising diffusion probabilistic models (DDPMs) \cite{ho2020denoisingdiffusionprobabilisticmodels}, denoising diffusion implicit models (DDIMs) \cite{song2021scorebasedgenerativemodelingstochastic, Song2021_ddim}, and Elucidated Diffusion Models (EDMs) \cite{karras2022elucidatingdesignspacediffusionbased}.

\subsection{Diffusion Training Objectives}

DDPMs corrupt a data sample \(x_0 \sim p_{\text{data}}\) over \(T\) steps with a variance schedule \(\{\beta_t\}_{t=1}^T\), which admits the closed-form marginal
\begin{equation}
x_t
=
\sqrt{\bar{\alpha}_t}\, x_0
+
\sqrt{1-\bar{\alpha}_t}\, \varepsilon,
\qquad
\varepsilon \sim \mathcal{N}(0,I),
\end{equation}
with \(\bar{\alpha}_t = \prod_{s=1}^t (1-\beta_s)\). A network trained to predict the added noise gives the standard DDPM objective
\begin{equation} \label{eq: denoising loss}
\mathcal{L}_{\text{DDPM}}
=
\mathbb{E}_{x_0,\,\varepsilon,\,t}
\left[
\left\|
\varepsilon - \varepsilon_\theta(x_t, t)
\right\|_2^2
\right],
\end{equation}
a reweighted variational bound on the negative log-likelihood. Every noise prediction induces a clean estimate
\begin{equation} \label{eqn: ddim sampling}
\widehat{x}_0
=
\left(x_t - \sqrt{1-\bar{\alpha}_t}\,\varepsilon_\theta(x_t,t)\right)/{\sqrt{\bar{\alpha}_t}},
\end{equation}
which DDIMs use to define a deterministic sampler with the same marginals as DDPMs. DDIMs keep the same training objective Eq.\ \eqref{eq: denoising loss} and change only the sampling procedure \citep{Song2021_ddim}. EDMs index corruption by the noise standard deviation \(\sigma\) instead of \(t\), with \(x_\sigma = x_0 + \sigma \varepsilon\), and train a denoiser \(D_\theta\) with the reweighted objective
\begin{equation}\label{eq:edm}
\mathcal{L}_{\text{EDM}}
=
\mathbb{E}_{x_0,\,\varepsilon,\,\sigma}
\left[
\lambda_{\text{EDM}}(\sigma)
\left\|
D_\theta(x_\sigma, \sigma) - x_0
\right\|_2^2
\right],
\end{equation}
where \(\lambda_{\text{EDM}}(\sigma)\) balances contributions across noise scales \citep{karras2022elucidatingdesignspacediffusionbased}. In all three formulations a clean prediction \(\widehat{x}_0\) is available at no extra cost, from Eq.\ \eqref{eqn: ddim sampling} or as \(D_\theta(x_\sigma,\sigma)\), and all three fit it with a pointwise squared error in the signal domain.

\subsection{The Spectral Allocation of the Error}

Let \(\mathcal{F}\) be the unitary discrete Fourier transform, computed via the fast Fourier transform (FFT) \citep{NumericalReceipes}, and write \(X = \mathcal{F}[x]\). By Parseval's identity we have that,
\begin{equation}
\label{eq:parseval_nd}
\|\widehat{x}_0 - x_0\|_2^2 = \textstyle\sum_{\omega} \big|\widehat{X}_0(\omega) - X_0(\omega)\big|^2.
\end{equation}
Thus, squared reconstruction error penalizes discrepancies in every Fourier bin with equal absolute weighting. However, omitting a low-power component incurs a small penalty even when its relative error is large. When signal power decreases with frequency, this can make missing high-frequency components inexpensive under the objective. Spectral regularization changes the representation or relative weighting of these discrepancies\citep{falck2025fourier, benita2025spectralanalysisdiffusionmodels, chen2025physiowavemultiscalewavelettransformerphysiological, ding2023patcheddenoisingdiffusionmodels, jiralerspong2025shapinginductivebiasdiffusion}.

However, the Fourier basis is global and lacks the spatial or temporal localization needed for non-stationary structure. Discrete wavelet transforms (DWTs) supply it through a hierarchical decomposition into approximation and detail coefficients at dyadic scales, with oriented sub-bands at each scale. Low-frequency coefficients encode coarse content, while high-frequency coefficients capture edges, textures, and transient features.

\section{Spectral Regularization} \label{sec:spectralreg}
We regularize the clean prediction $\hat{x}_0$ with $L^1$ discrepancies in the Fourier and wavelet domains to change the objective’s sensitivity to spectral magnitudes and localized multiscale structure.

\subsection{Fourier-Regularized Diffusion Models}

Let \(x_t\) be a noisy version of \(x_0\) at diffusion time \(t\), and let \(\widehat{x}_0 = \widehat{x}_\theta(x_t, t)\) be the predicted clean sample. We express the Fourier transform in polar form,
\begin{equation}
\label{eq:fourier_polar}
\mathcal{F}[x](\omega) = A(\omega) \, \exp\!\bigl(i\,\phi(\omega)\bigr),
\end{equation}
with amplitude \(A = |\mathcal{F}[x]|\) and phase \(\phi\), and denote these by \(A_0, \phi_0\) for \(x_0\) and by \(\widehat{A}_0, \widehat{\phi}_0\) for \(\widehat{x}_0\).

\paragraph{Amplitude Loss.}
The first regularizer enforces agreement between the amplitude spectra of the generated and target samples,
\begin{equation}
\label{eq:fourier_amp_loss}
\mathcal{L}_{\mathrm{F}}^{\mathrm{A}} = \mathbb{E}_{x_0,\,t}\left[\left\|A_0 - \widehat{A}_{0} \right\|_{1} \right].
\end{equation}
The amplitude spectrum captures how signal energy is distributed across frequencies, independent of spatial alignment. The amplitude spectrum describes the magnitude of each frequency component and is invariant to circular translations Thus, it measures spectral discrepancies without directly penalizing phase differences.

\paragraph{Amplitude-and-Phase Loss.}
Amplitude fixes how much energy each band carries, while phase fixes where the corresponding structure sits. The second regularizer incorporates both,
\begin{equation}\label{eq:fourier_amp_scale_loss}
\mathcal{L}_{\mathrm{F}}^{\mathrm{AP}} = \mathbb{E}_{x_0,\,t}\left[\left\|A_0 - \widehat{A}_{0} \right\|_{1} \left( 1+ \left\|\phi_0 - \widehat{\phi}_{0}\right\|_{1}\right)\right].
\end{equation}
Phase disagreement acts as a multiplicative weight on the amplitude discrepancy. At differentiable points, the phase-gradient contribution is scaled by $|A_0-\hat A_0|_1$. The loss vanishes whenever the amplitude spectra match, regardless of phase. This weighting does not remove principal-phase discontinuities or sensitivity near zero Fourier coefficients.

\subsection{Wavelet-Regularized Diffusion Models}

We use  \(W_0^{(s,\ell)}(b)\) and \(\widehat{W}_0^{(s,\ell)}(b)\) to denote the discrete wavelet coefficients of \(x_0\) and \(\widehat{x}_0\), where \(s\) indexes scale, \(\ell\) indexes the orientation or sub-band, and \(b\) denotes spatial or temporal location. The wavelet loss enforces alignment between the coefficients at all scales and orientations,
\begin{equation}
\label{eq:wavelet_coeff_loss}
\mathcal{L}_{\mathrm{W}}=\mathbb{E}_{x_0,\,t}\left[\sum_{s,\ell}\gamma_{s,\ell}\left\|W_0^{(s,\ell)}-\widehat{W}_0^{(s,\ell)}\right\|_{1}\right],
\end{equation}
where the norm is taken over locations \(b\) and \(\gamma_{s,\ell}\) weights each scale and sub-band. It encourages the model to match localized features such as edges, textures, and transient events at every resolution level.

\subsection{Training Objective}

The final training objective augments the standard diffusion loss with a spectral regularizer,
\begin{equation}
\label{eq:fourier_diffusion_objective}
\mathcal{L}_{\text{total}}=\mathcal{L}+\lambda\,\mathcal{L}_{\mathrm{S}},
\end{equation}
where \(\mathcal{L}\) is the denoising objective Eq.\ \eqref{eq: denoising loss} or Eq.\ \eqref{eq:edm}, \(\mathcal{L}_{\mathrm{S}}\) is \(\mathcal{L}_{\mathrm{F}}^{\mathrm{A}}\), \(\mathcal{L}_{\mathrm{F}}^{\mathrm{AP}}\), or \(\mathcal{L}_{\mathrm{W}}\), and \(\lambda\) controls the strength of the regularization. Only the loss changes, so the forward process, architecture, and sampler are untouched, and the overhead is only a single FFT/DWT of \(\widehat{x}_0\) and \(x_0\) per training step. The spectral term is evaluated on \(\widehat{x}_0\), which each formulation obtains from the network output with a different gain.

\paragraph{Spectral Regularization for Distillation.}
Because only the loss changes, the spectral loss regularizers can also be applied to objectives other than denoising. In few-step distillation, the student must reproduce a many-step trajectory
of the teacher in a few large steps and tends to lose fine-scale structure. The
student is trained to match, at each of its time points $\sigma_n$, the state
$x^{\mathrm{T}}_{\sigma_n}$ of the teacher's trajectory from the same noise.
Intermediate states are dominated by the noise realization, so the spectral term acts
only at the final time point, where both states are essentially clean,
\begin{equation}
  \mathcal{L}_{\mathrm{distill}} = \sum_{n} \big\| x^{\mathrm{S}}_{\sigma_{n+1}} - x^{\mathrm{T}}_{\sigma_{n+1}} \big\|_1
  + \lambda\, \mathcal{S}\big( x^{\mathrm{S}}_{\sigma_{\min}}, x^{\mathrm{T}}_{\sigma_{\min}} \big),
  \label{eq:distill-obj}
\end{equation}
where $\mathcal{S}$ is the per-sample discrepancy $\ell_{\mathrm{S}}$ with the teacher's
state in place of $x_0$.

\subsection{Spectral Regularization on Graphs}
\label{sec:graph-loss}

The regularizers need only a transform of $\widehat{x}_0$ whose coefficients
measure structure that the squared error leaves unconstrained, so they are not
limited to grids. On a graph with symmetric weighted adjacency matrix $A$, the
eigenvectors of the Laplacian $L(A) = \mathrm{diag}(A\mathbf{1}) - A$ form the
graph Fourier basis and its eigenvalues play the role of frequencies
\citep{shuman2013emerging}. The sorted eigenvalues do not depend on the order of
the nodes, just as the Fourier amplitude is invariant under translation. For a
diffusion model that denoises adjacency matrices, let $\widehat{A}_0$ be the
weighted adjacency of the clean prediction. The spectral eigenvalue loss is given by
\begin{equation}
  \mathcal{L}_{\mathrm{G}} = \mathbb{E}_{x_0,\,t}\Big[\frac{1}{n}\sum_{i=1}^{n}
  \big|\lambda_i\big(L(A_0)\big) - \lambda_i\big(L(\widehat{A}_0)\big)\big|\Big],
  \label{eq:graph_eig_loss}
\end{equation}
with the eigenvalues of each $n$-node graph in ascending order. It enters
Eq.\ \eqref{eq:fourier_diffusion_objective} as $\mathcal{L}_{\mathrm{S}}$, at the cost of
one eigendecomposition of an $n \times n$ symmetric matrix per graph.

\section{Experiments} \label{sec:experiments}
We report fine-tuning experiments for audio and graph models. We also report distillation results for image data. Results for fine-tuning image models are reported in Appendix~\ref{sec:image-exps}. Our fine-tuning experiments use released pretrained checkpoints as baselines. The distillation experiments include matched baselines trained with $\lambda=0$.



\subsection{Audio Dataset}
\begin{table*}
  \caption{Audio generation quality metrics for DiffWave fine-tuning with spectral regularization.
FAD measures distributional similarity in audio embedding space (lower is better),
UTMOS estimates perceptual naturalness (higher is better),
PESQ measures perceptual speech quality (higher is better),
MR-STFT measures multi-resolution spectral error (lower is better),
and NDB evaluates distributional coverage and mode balance (lower is better). We report the average and standard deviation of five runs with different seeds.}
  \label{tab:audio_metrics}
  \centering
  \begin{footnotesize}
  \begin{sc}
  \begin{tabular}{cccccccc}
    \toprule
    Method &
    $\lambda$ &
    \shortstack{FAD $\downarrow$} &
    \shortstack{UTMOS $\uparrow$} &
    \shortstack{PESQ $\uparrow$} &
    \shortstack{MR-STFT $\downarrow$} &
    \shortstack{NDB $\downarrow$} \\
    \midrule
    Diffwave & 0 &
    1.994$\pm$0.008 &
    3.941$\pm$0.005 &
    3.440$\pm$0.002 &
    1.217$\pm$0.001 &
    0.63$\pm$0.02 \\
    \midrule

    \multirow{3}{*}{Amp} &
    $10^{-4}$ &
    \bf{1.462$\pm$0.006} &
    3.953$\pm$0.009 &
    3.477$\pm$0.003 &
    1.1802$\pm$0.0005 &
    0.65$\pm$0.03 \\
     &
    $10^{-5}$ &
    1.609$\pm$0.007 &
    3.953$\pm$0.009 &
    3.476$\pm$0.001 &
    1.1930$\pm$0.0004 &
    0.63$\pm$0.02 \\
     &
    $10^{-6}$ &
    1.775$\pm$0.008 &
    3.952$\pm$0.006 &
    3.491$\pm$0.003 &
    1.1958$\pm$0.0005 &
    0.64$\pm$0.03 \\
    \midrule

    \multirow{3}{*}{Amp+Phase} &
    $10^{-4}$ &
    1.694$\pm$0.009 &
    3.969$\pm$0.008 &
    \bf{3.516$\pm$0.002} &
    1.1896$\pm$0.0003 &
    0.66$\pm$0.02 \\
     &
    $10^{-5}$ &
    1.543$\pm$0.012 &
    \bf{3.988$\pm$0.003} &
    3.495$\pm$0.002 &
    1.1773$\pm$0.0004 &
    \bf{0.59$\pm$0.01} \\
     &
    $10^{-6}$ &
    1.539$\pm$0.007 &
    3.976$\pm$0.008 &
    3.344$\pm$0.003 &
    1.1921$\pm$0.0003 &
    0.65$\pm$0.02 \\
    \midrule

    \multirow{3}{*}{Haar} &
    $10^{-4}$ &
    1.729$\pm$0.016 &
    3.965$\pm$0.006 &
    3.466$\pm$0.003 &
    \bf{1.1708$\pm$0.0006} &
    \bf{0.59$\pm$0.02} \\
     &
    $10^{-5}$ &
    1.992$\pm$0.014 &
    3.988$\pm$0.008 &
    3.485$\pm$0.003 &
    1.2163$\pm$0.0002 &
    0.66$\pm$0.02 \\
     &
    $10^{-6}$ &
    2.123$\pm$0.010 &
    3.923$\pm$0.002 &
    3.359$\pm$0.002 &
    1.2343$\pm$0.0002 &
    0.69$\pm$0.03 \\
    \midrule

    \multirow{3}{*}{Bior} &
    $10^{-4}$ &
    1.492$\pm$0.011 &
    3.977$\pm$0.009 &
    3.500$\pm$0.002 &
    1.1768$\pm$0.0002 &
    0.62$\pm$0.01 \\
     &
    $10^{-5}$ &
    2.649$\pm$0.011 &
    3.927$\pm$0.007 &
    3.303$\pm$0.001 &
    1.2949$\pm$0.0004 &
    0.67$\pm$0.04 \\
     &
    $10^{-6}$ &
    1.520$\pm$0.008 &
    3.985$\pm$0.008 &
    3.466$\pm$0.003 &
    1.1787$\pm$0.0003 &
    0.64$\pm$0.02 \\

    \bottomrule
  \end{tabular}
  \end{sc}
  \end{footnotesize}
  \vskip -0.1in
\end{table*}
We additionally evaluate spectral regularization for audio generation by fine-tuning a pretrained DiffWave model using the same loss-level protocol as in our image experiments \cite{kong2021diffwaveversatilediffusionmodel}. Specifically, we optimize the standard DiffWave DDPM denoising objective and augment it with one of our proposed spectral losses, namely Fourier amplitude, Fourier amplitude-phase, and wavelet-based regularizers. We fine-tune the official implementation for 150,000 steps on the LJSpeech-1.1 dataset starting from the publicly released pretrained checkpoint using different constant $\lambda$ values ranging from $10^{-4}$ to $10^{-6}$ on the Fourier and wavelet (Haar and bi-orthogonal) losses ~\cite{ljspeech}. We also fine-tune the the model with its usual training loss (i.e.\ $\lambda = 0$) as a baseline. The results are summarized in Table~\ref{tab:audio_metrics}. Additional training details are presented in Appendix~\ref{app:training_details_diffwave}.

Overall, we find that the spectral fine-tuning improves the baseline DiffWave model across  perceptual, spectral, and distributional metrics, despite being applied as a lightweight fine-tuning procedure. However the exact combination of spectral loss and regularizer weight makes a substantial difference.  All spectral losses outperform the DiffWave baseline in FAD and PESQ for certain choice of $\lambda$, indicating that explicit spectral-domain biasing effectively corrects spectral mismatches that are weakly constrained by time-domain denoising alone. Fourier amplitude regularization yields the strongest FAD improvements, achieving the best overall score at moderate regularization strength, suggesting that matching global magnitude statistics is sufficient to recover dominant spectral structure that drives perceptual distance. In contrast, the amplitude-phase loss produces the most balanced gains across metrics, attaining the highest UTMOS and PESQ values and the lowest NDB. The amplitude–phase variant is competitive on UTMOS and PESQ, while amplitude alone achieves the lowest FAD, indicating that the preferred formulation depends on the evaluation metric.

Wavelet-based regularization exhibits complementary behavior. The Haar wavelet loss achieve the lowest MR-STFT distance at higher $\lambda$, highlighting improved multi-resolution temporal coherence. The bi-orthognal results vary substantially with the regularization weight. Overall, no single spectral loss dominates across all criteria. Thus, spectral regularization acts a controllable inductive bias whose effect depends on both representation choice and loss weighting.

\subsection{Graph Dataset}
\label{sec:graph}
\begin{table*}[t]
  \caption{Graph generation on QM9 with SwinGNN fine-tuned from the released
  checkpoint with the eigenvalue loss at three weights $\lambda$, for samplers
  with 16 to 256 steps. Invalid counts the molecules per
  $10^4$ that fail the valence check before correction. Unique and Novel are the
  percentages of unique molecules and of molecules not in the training set. FCD
  and NSPDK MMD ($\times 10^4$) are computed against the test set (lower is better). We report the mean and standard deviation over 3 sampling/training seeds with 10k molecules per evaluation.}
  \label{tab:graph}
  \centering
  \begin{footnotesize}
  \begin{sc}
  \setlength{\tabcolsep}{4pt}
  \begin{tabular}{cccccccc}
  \toprule
  Steps & Method & $\lambda$ & Invalid $\downarrow$ & Unique $\uparrow$ & Novel & FCD $\downarrow$ & NSPDK $\downarrow$ \\
  \midrule
  \multirow{4}{*}{16} & SwinGNN & -- & $76.7 \pm 14.3$ & $95.2 \pm 0.2$ & $30.3 \pm 0.5$ & $0.623 \pm 0.017$ & $28.6 \pm 1.2$ \\
   & \multirow{3}{*}{Eigenvalue Loss} & 0.018 & {\boldmath$76.3 \pm 13.7$} & $95.2 \pm 0.2$ & $31.6 \pm 0.5$ & $0.640 \pm 0.015$ & $31.0 \pm 1.5$ \\
   &  & 0.059 & $84.9 \pm 14.1$ & {\boldmath$95.3 \pm 0.2$} & $31.8 \pm 0.4$ & {\boldmath$0.609 \pm 0.016$} & $33.1 \pm 1.4$ \\
   &  & 0.176 & $100.2 \pm 12.1$ & $95.2 \pm 0.1$ & $31.9 \pm 0.8$ & {\boldmath$0.615 \pm 0.011$} & $39.1 \pm 1.2$ \\
  \midrule
  \multirow{4}{*}{32} & SwinGNN & -- & $46.0 \pm 4.6$ & $95.7 \pm 0.3$ & $21.4 \pm 0.7$ & $0.306 \pm 0.021$ & $14.1 \pm 1.5$ \\
   & \multirow{3}{*}{Eigenvalue Loss} & 0.018 & $50.8 \pm 6.0$ & $95.7 \pm 0.2$ & $22.2 \pm 0.6$ & {\boldmath$0.305 \pm 0.014$} & $15.0 \pm 1.6$ \\
   &  & 0.059 & $51.8 \pm 6.6$ & {\boldmath$95.8 \pm 0.1$} & $22.2 \pm 0.7$ & {\boldmath$0.286 \pm 0.010$} & $16.1 \pm 1.7$ \\
   &  & 0.176 & $49.9 \pm 8.1$ & {\boldmath$95.9 \pm 0.2$} & $22.1 \pm 0.8$ & {\boldmath$0.290 \pm 0.005$} & $19.5 \pm 1.7$ \\
  \midrule
  \multirow{4}{*}{64} & SwinGNN & -- & $20.7 \pm 2.3$ & $96.2 \pm 0.3$ & $15.7 \pm 0.2$ & $0.185 \pm 0.011$ & $7.8 \pm 0.5$ \\
   & \multirow{3}{*}{Eigenvalue Loss} & 0.018 & $26.6 \pm 2.7$ & $96.0 \pm 0.2$ & $16.4 \pm 0.4$ & {\boldmath$0.183 \pm 0.004$} & $8.5 \pm 0.6$ \\
   &  & 0.059 & $26.3 \pm 4.2$ & $96.0 \pm 0.2$ & $16.3 \pm 0.4$ & {\boldmath$0.170 \pm 0.003$} & $8.8 \pm 0.5$ \\
   &  & 0.176 & $25.9 \pm 4.8$ & $96.1 \pm 0.2$ & $16.3 \pm 0.4$ & {\boldmath$0.178 \pm 0.005$} & $10.6 \pm 0.9$ \\
  \midrule
  \multirow{4}{*}{256} & SwinGNN & -- & $10.7 \pm 4.7$ & $96.2 \pm 0.3$ & $12.1 \pm 0.3$ & $0.102 \pm 0.001$ & $2.7 \pm 0.3$ \\
   & \multirow{3}{*}{Eigenvalue Loss} & 0.018 & $14.3 \pm 3.9$ & $96.2 \pm 0.2$ & $12.8 \pm 0.4$ & $0.104 \pm 0.002$ & $2.9 \pm 0.2$ \\
   &  & 0.059 & $13.1 \pm 3.2$ & $96.2 \pm 0.2$ & $12.8 \pm 0.4$ & $0.108 \pm 0.002$ & $3.2 \pm 0.2$ \\
   &  & 0.176 & $13.2 \pm 3.9$ & $96.2 \pm 0.2$ & $12.8 \pm 0.4$ & $0.149 \pm 0.006$ & $4.4 \pm 0.2$ \\
  \bottomrule
  \end{tabular}
  \end{sc}
  \end{footnotesize}
  \vskip -0.1in
\end{table*}
We now test our spectral regularization framework on graph generation, where the
eigenvalue loss acts as the spectral regularizer. We fine-tune SwinGNN, an EDM model that denoises the adjacency matrices of molecules, from its released checkpoint on the QM9 molecular dataset for 500 epochs \citep{yan2024swingnn, ramakrishnan2014qm9}. We add the eigenvalue loss on the predicted bond orders to the adjacency term of its objective. We use three weights $\lambda$, chosen so that the spectral term is $3\%$, $10\%$ and $30\%$ of the adjacency loss at the start of fine-tuning, and three training seeds per weight (see Appendix~\ref{app:graph} for details).

We sample 10k molecules per evaluation with the stochastic Heun sampler of
SwinGNN, using 16, 32 and 64 steps and the default 256. Following \citet{jo2022score}, we report the number of invalid molecules before valence correction, uniqueness and novelty, and FCD
\citep{preuer2018fcd} and NSPDK MMD \citep{costa2010nspdk} against the test set. The released checkpoint is evaluated with the same EMA rate as the fine-tunes, $0.999$, since the EMA rate alone changes its number of invalid molecules by up to a factor of $2.5$ (see Appendix~\ref{app:graph}).

Table~\ref{tab:graph} shows a trade-off that depends on the sampling budget. With 16 to 64 steps the middle weight $\lambda = 0.059$ gives the lowest FCD at every budget ($0.286$ against $0.306$ at 32 steps), although over three sampling seeds these gains are not significant. At the default 256 steps, where the released model is already close to the FCD of $0.077$ that 10k training molecules score against the test set (Appendix~\ref{app:graph}), every weight raises FCD, to $0.104$, $0.108$ and $0.149$. NSPDK, in contrast, grows with the weight at every budget. Novelty and, with 32 or more steps, the number of invalid molecules shift by about the same amount at every weight, which points to fine-tuning itself and not to the spectral term.

FCD and NSPDK can disagree because they measure different properties. FCD is a Fr\'echet distance between ChemNet activations, which are features learned to predict biological activity, and so reflects the chemical and biological similarity of the generated and test molecules. NSPDK-MMD compares the distributions of small labeled subgraphs, and so reflects exact local topology. A change in the samples can therefore lower one metric and raise the other. Here the eigenvalue loss moves the local substructure statistics away from the test set at every budget, while its effect on FCD depends on the sampling budget.

\subsection{Few-Step Distillation} \label{sec:distillation}
%
\begin{figure}[t]
\centering
\includegraphics[width=\linewidth]{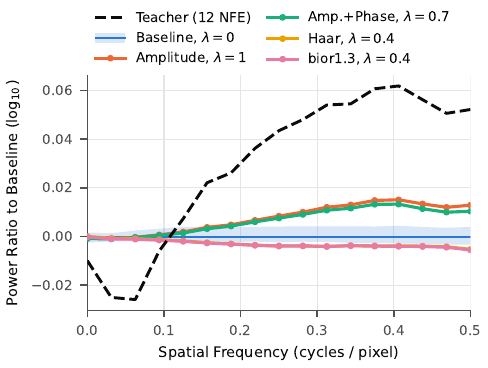}
\caption{Radially averaged power of 2-NFE CIFAR-10 students relative to the
$\lambda = 0$ student ($\log_{10}$ ratio, shaded band is the range over three
baseline seeds) for each loss at its middle weight, averaged over its seeds with each seed paired with the baseline of the same seed, and for the 12-NFE teacher. The Fourier losses move power toward the teacher above about $0.1$ cycles
per pixel, and the wavelet losses move it away.}
\label{fig:distill_spectra}
\end{figure}

Next we apply our spectral diffusion framework to distillation. We distill with the objective of Eq.~\eqref{eq:distill-obj}, using the code and settings of Simple and Fast Distillation (SFD) \citep{zhou2024sfd, zhou2024amed}. A 2-NFE student initialized from a pretrained EDM VP model (three Euler steps, the first analytical learns to match the states of a 12-NFE DPM-Solver++ trajectory of the teacher from the same noise \citep{lu2022dpmsolverpp}. See Appendix~\ref{app:distill} for more details.

We distill on unconditional CIFAR-10 and on FFHQ $64 \times 64$ and score each student
by FID on 50k samples. Each loss is run at three weights, chosen so that the spectral term is roughly $10\%$, $33\%$ and $100\%$ of the pixel term (see Appendix~\ref{app:distill-calib}), and each regularized student is compared with the baseline student trained and sampled from the same seed.

\begin{table}[t]
\centering
\caption{Distillation: FID of 2-NFE SFD students on CIFAR-10 ($32\times32$) and FFHQ
($64 \times 64$) over 3 paired seeds. Bold marks the lowest FID per dataset.}
\label{tab:distill}
\small
\setlength{\tabcolsep}{3.7pt}
\begin{tabular}{lrcc}
\toprule
 & & \multicolumn{2}{c}{FID $\downarrow$} \\
\cmidrule(lr){3-4}
Loss & $\lambda$ & CIFAR-10 & FFHQ \\
\midrule
SFD Baseline    & 0    & $4.475 \pm 0.020$ & $5.837 \pm 0.016$ \\
\addlinespace
Amplitude   & 0.3  & $4.447 \pm 0.038$ & $5.777 \pm 0.032$ \\
            & 1    & {\boldmath$4.423 \pm 0.041$} & $5.733 \pm 0.051$ \\
            & 3.5  & $4.500 \pm 0.039$ & {\boldmath$5.712 \pm 0.053$} \\
\addlinespace
Amp.+phase  & 0.2  & $4.448 \pm 0.043$ & $5.781 \pm 0.032$ \\
            & 0.7  & $4.489 \pm 0.038$ & $5.730 \pm 0.050$ \\
            & 2    & $4.626 \pm 0.036$ & $5.731 \pm 0.042$ \\
\addlinespace
Haar        & 0.13 & $4.558 \pm 0.034$ & $5.869 \pm 0.039$ \\
            & 0.4  & $4.665 \pm 0.030$ & $5.919 \pm 0.061$ \\
            & 1.3  & $4.865 \pm 0.041$ & $6.001 \pm 0.097$ \\
\addlinespace
bior1.3     & 0.12 & $4.557 \pm 0.024$ & $5.864 \pm 0.034$ \\
            & 0.4  & $4.675 \pm 0.033$ & $5.918 \pm 0.056$ \\
            & 1.2  & $4.852 \pm 0.036$ & $5.986 \pm 0.091$ \\
\bottomrule
\end{tabular}
\end{table}

Table~\ref{tab:distill} shows that the Fourier amplitude loss improves the 2-NFE
student on both datasets as the FID score decreases by 0.052 (1\%) and 0.125 (2\%) for CIFAR-10 and FFHQ respectively in the best settings. Compared with amplitude alone, amplitude with phase yields smaller FID improvements on CIFAR-10 and similar improvements on FFHQ. In contrast, both wavelet losses degrade the student performance at every weight on both datasets. Because their weights span the same relative sizes as those of the amplitude loss, the amplitude gain is not a generic effect of adding a loss term.

Figure~\ref{fig:distill_spectra} shows how the losses change the spectrum of the
student's samples. The baseline student has less high-frequency power than its teacher.
The Fourier losses raise this power toward the teacher's, while the wavelet losses
lower it further. A closer average spectrum does not by itself mean better samples,
however. On CIFAR-10, the Fourier losses continue to bring high-frequency power closer to the teacher’s as $\lambda$ grows while FID worsens at the larger weights. On FFHQ, the same weights change the spectrum by about as much while FID does not worsen up to the
largest weight tested (Appendix~\ref{app:distill-ffhq}). The weight must therefore be
chosen by sample quality on the target data. Appendix~\ref{app:distill-discussion}
discusses why the wavelet result differs from fine-tuning.

\section{Conclusion}
We introduced a spectral regularization framework for diffusion models that adds penalties on the Fourier amplitude/phase, the wavelet coefficients, or the graph Laplacian eigenvalues of the clean prediction, while leaving the sampler and architecture unchanged. We show improvements in fine-tuning and distillation tasks across image, audio, and graph domains. In distillation, the Fourier amplitude loss improves few-step students while the wavelet losses degrade them, and in audio fine-tuning different losses favor different metrics. For graphs, the effect of the eigenvalue loss varies across evaluation metrics and sampling budgets.


\bibliography{paper_bib}

\clearpage
\appendix
\thispagestyle{empty}

\onecolumn
\aistatstitle{Appendix}

\section{Fine-tuning on Image Datasets} \label{sec:image-exps}

We study spectral regularization as a lightweight fine-tuning strategy applied to pretrained EDM models \cite{karras2022elucidatingdesignspacediffusionbased}. For each dataset and EDM formulation, models are fine-tuned for 5 optimization steps using the standard EDM denoising objective augmented with a spectral loss. This setup isolates the effect of loss-level spectral biasing without modifying the model architecture, diffusion formulation, or sampler. More training details are in Appendix~\ref{app:training_details_edm}.

We consider both variance-preserving (VP) and variance-exploding (VE) EDM variants. For each, we evaluate four spectral losses: (i) Fourier amplitude, (ii) Fourier amplitude+phase, (iii) Haar wavelet, and (iv) bi-orthogonal 1.3 (bior13) wavelet regularization. Fourier transforms are computed using PyTorch FFTs, while wavelet transforms are implemented based on PyWavelets \cite{Lee2019}. Experiments are conducted on CIFAR-10 ($32\times32$), FFHQ, and AFHQv2 ($64\times64$), following the standard EDM evaluation protocol. CIFAR-10 is evaluated under conditional sampling, whereas FFHQ and AFHQv2 are evaluated unconditionally. We consider two choices for the regularization weight. The first being the ``weighted'' setting where $\lambda = \lambda_{\mathrm{EDM}}$ (see Eq.~\ref{eq:edm}), and the ``unweighted'' setting where $\lambda=1$. Generative quality is measured using Fr\'echet Inception Distance (FID), computed between 50,000 generated samples and the full real dataset. Results are averaged over three random seeds. Some sampled images for the FFHQ and AFHQ datasets are shown in Appendix \ref{app: gen_image_afhq_ffhq}.

Table \ref{tab:image_fid} reports FID scores for CIFAR-10, AFHQ, and FFHQ under both VE- and VP-EDM formulations. Since all experiments start from strong pretrained EDM baselines, the scope for improvement is necessarily limited. On CIFAR-10 (conditional), spectral losses have negligible effect, with all methods performing within the standard deviation of the EDM baseline, indicating limited benefit when conditional structure is already well captured.

On the higher-resolution AFHQ and FFHQ datasets, we observe small but reliable FID reductions (typically $0.02-0.07$) across multiple spectral losses and EDM variants, with no cases of systematic degradation. These gains are comparable in magnitude across AFHQ and FFHQ, indicating that the proposed losses behave similarly on distinct but equally challenging natural image distributions.

Importantly, the improvements are achieved with only a handful of fine-tuning steps and without modifying the architecture or sampler, highlighting that spectral regularization acts as a stable and data-efficient bias rather than an aggressive optimization mechanism.

Overall, amplitude-phase regularization is the most consistently competitive method, achieving the best or tied-best performance on FFHQ and remaining close to optimal elsewhere. These results suggest that spectral regularization is most effective in higher-resolution, unconditional settings where diffusion models struggle to capture fine-scale structure, and offer limited benefits when baseline performance is already near saturation.

\begin{table*}

  \caption{Fréchet Inception Distance (FID) scores for different spectral regularizers and EDM variants. All models are fine-tuned from pretrained EDM baselines. Lower values indicate better generative performance.}
  \label{tab:image_fid}
  \centering
    \begin{footnotesize}
      \begin{sc}
        \begin{tabular}{cccccccc}
          \toprule
          Dataset & \shortstack{EDM\\variant} & \shortstack{EDM Sampling\\Steps} & Conditioning & Regularizer & \shortstack{Weighted\\FID} &  \shortstack{Unweighted\\FID} &  \shortstack{EDM\\FID} \\
          \midrule
          \multirow{4}{*}{CIFAR} & \multirow{4}{*}{VE} & \multirow{4}{*}{18} & \multirow{4}{*}{Cond} & Amp & 1.82$\pm$0.01 & 1.82$\pm$0.01 & \multirow{4}{*}{1.81$\pm$0.01} \\
           &  &  &  & Amp Phase & 1.82$\pm$0.01 & 1.82$\pm$0.01 &  \\
           &  &  &  & Haar & \bf{1.81$\pm$0.01} & \bf{1.81$\pm$0.02} &  \\
           &  &  &  & Bior13 & 1.82$\pm$0.01 & \bf{1.81$\pm$0.02} &  \\
          \midrule
          \multirow{4}{*}{CIFAR} & \multirow{4}{*}{VP} & \multirow{4}{*}{18} & \multirow{4}{*}{Cond} & Amp & 1.84$\pm$0.02 & \bf{1.84$\pm$0.02} & \multirow{4}{*}{1.84$\pm$0.02} \\
           &  &  &  & Amp Phase & 1.84$\pm$0.02 & \bf{1.84$\pm$0.02}  &  \\
           &  &  &  & Haar & 1.84$\pm$0.02 & \bf{1.84$\pm$0.02}  &  \\
           &  &  &  & Bior13 & \bf{1.83$\pm$0.02} & \bf{1.84$\pm$0.02}  &  \\
          \midrule
          \multirow{4}{*}{AFHQ} & \multirow{4}{*}{VE} & \multirow{4}{*}{40} & \multirow{4}{*}{Uncond} & Amp & \bf{2.13$\pm$0.00} & \bf{2.14$\pm$0.00} & \multirow{4}{*}{2.17$\pm$0.00} \\
           &  &  &  & Amp Phase & \bf{2.13$\pm$0.01} & 2.16$\pm$0.01 &  \\
           &  &  &  & Haar & 2.14$\pm$0.01 & \bf{2.14$\pm$0.00} &  \\
           &  &  &  & Bior13 & 2.14$\pm$0.01 & 2.14$\pm$0.01 &  \\
          \midrule
          \multirow{4}{*}{AFHQ} & \multirow{4}{*}{VP} & \multirow{4}{*}{40} & \multirow{4}{*}{Uncond} & Amp & \bf{2.03$\pm$0.02} & 2.05$\pm$0.02 & \multirow{4}{*}{2.04$\pm$0.00} \\
           &  &  &  & Amp Phase & 2.04$\pm$0.02 & \bf{2.03$\pm$0.02} &  \\
           &  &  &  & Haar & 2.05$\pm$0.02 & \bf{2.03$\pm$0.03} &  \\
           &  &  &  & Bior13 & 2.07$\pm$0.02 & 2.05$\pm$0.02 &  \\
          \midrule
          \multirow{4}{*}{FFHQ} & \multirow{4}{*}{VE} & \multirow{4}{*}{40} & \multirow{4}{*}{Uncond} & Amp & 2.5$\pm$0.02 & 2.51$\pm$0.02 & \multirow{4}{*}{2.56$\pm$0.03} \\
           &  &  &  & Amp Phase & \bf{2.49$\pm$0.02} & 2.50$\pm$0.02 &  \\
           &  &  &  & Haar & 2.5$\pm$0.02 & \bf{2.49$\pm$0.02} & \\
           &  &  &  & Bior13 & 2.51$\pm$0.02 & 2.5$\pm$0.02 &  \\
          \midrule
          \multirow{4}{*}{FFHQ} & \multirow{4}{*}{VP} & \multirow{4}{*}{40} & \multirow{4}{*}{Uncond} & Amp & 2.35$\pm$0.03 & \bf{2.31$\pm$0.03} & \multirow{4}{*}{2.38$\pm$0.01} \\
           &  &  &  & Amp Phase & \bf{2.33$\pm$0.03} & 2.34$\pm$0.03 &  \\
           &  &  &  & Haar & \bf{2.33$\pm$0.04} & 2.33$\pm$0.04 &  \\
           &  &  &  & Bior13 & 2.34$\pm$0.04 & 2.32$\pm$0.04 &  \\
          \bottomrule
        \end{tabular}
      \end{sc}
    \end{footnotesize}
  \vskip -0.1in
\end{table*}

\section{Training Details for Image and Audio Fine Tuning} \label{app:training_details}
\subsection{EDM Configuration}
\label{app:training_details_edm}
We adopt the EDM framework of \citep{karras2022elucidatingdesignspacediffusionbased} as our pretrained diffusion backbone. We fine-tune eight publicly released checkpoints corresponding to the continuous-time DDPM++ (VP) and NCSN++ (VE) variants, pretrained on CIFAR-10 (conditional and unconditional), FFHQ-64 (unconditional), and AFHQv2 (unconditional). The pretrained weights are obtained from the official release and correspond to the following datasets and variants:
\begin{itemize}
    \item \textbf{AFHQv2-64 (unconditional)}
    \begin{itemize}
        \item \textbf{VE}:
        {\footnotesize\url{https://nvlabs-fi-cdn.nvidia.com/edm/pretrained/edm-afhqv2-64x64-uncond-ve.pkl}}
        \item \textbf{VP}:
        {\footnotesize\url{https://nvlabs-fi-cdn.nvidia.com/edm/pretrained/edm-afhqv2-64x64-uncond-vp.pkl}}
    \end{itemize}

    \item \textbf{CIFAR-10 (conditional)}
    \begin{itemize}
        \item \textbf{VE}:
        {\footnotesize\url{https://nvlabs-fi-cdn.nvidia.com/edm/pretrained/edm-cifar10-32x32-cond-ve.pkl}}
        \item \textbf{VP}:
        {\footnotesize\url{https://nvlabs-fi-cdn.nvidia.com/edm/pretrained/edm-cifar10-32x32-cond-vp.pkl}}
    \end{itemize}

    \item \textbf{CIFAR-10 (unconditional)}
    \begin{itemize}
        \item \textbf{VE}:
        {\footnotesize\url{https://nvlabs-fi-cdn.nvidia.com/edm/pretrained/edm-cifar10-32x32-uncond-ve.pkl}}
        \item \textbf{VP}:
        {\footnotesize\url{https://nvlabs-fi-cdn.nvidia.com/edm/pretrained/edm-cifar10-32x32-uncond-vp.pkl}}
    \end{itemize}

    \item \textbf{FFHQ-64 (unconditional)}
    \begin{itemize}
        \item \textbf{VE}:
        {\footnotesize\url{https://nvlabs-fi-cdn.nvidia.com/edm/pretrained/edm-ffhq-64x64-uncond-ve.pkl}}
        \item \textbf{VP}:
        {\footnotesize\url{https://nvlabs-fi-cdn.nvidia.com/edm/pretrained/edm-ffhq-64x64-uncond-vp.pkl}}
    \end{itemize}
\end{itemize}

All models are fine-tuned using the same optimization protocol as in the original EDM work. We use a per-GPU batch size of 16 and run experiments on NVIDIA A4000 and A6000 GPUs. Dataset-specific hyperparameters for fine-tuning are summarized in Table~\ref{tab:configuration}.

\begin{table}[t]
\centering
\small
\setlength{\tabcolsep}{6pt}
\renewcommand{\arraystretch}{1.25}
\caption{EDM fine-tuning hyperparameters used for all experiments.}
\label{tab:configuration}
\begin{tabular}{lcccccc}
\toprule
\textbf{Dataset} &
\textbf{Duration} &
\textbf{Batch} &
\textbf{LR} &
\textbf{CRes} &
\textbf{Dropout} &
\textbf{Augment} \\
\midrule
CIFAR-10 (cond/uncond) & 0.5 & 16 & $2{\times}10^{-4}$ & -- & -- & -- \\
AFHQ-64 (uncond)      & 0.5 & 16  & $5{\times}10^{-5}$ & 1,2,2,2 & 0.25 & 0.15 \\
FFHQ-64 (uncond)      & --  & 16  & $5{\times}10^{-5}$ & 1,2,2,2 & 0.05 & 0.15 \\
\bottomrule
\end{tabular}
\end{table}

\subsection{Diffwave Configuration}\label{app:training_details_diffwave}

We adopt DiffWave \citep{kong2021diffwaveversatilediffusionmodel} as our pretrained audio diffusion backbone and fine-tune the official checkpoint released by LMNT.\footnote{\url{https://github.com/lmnt-com/diffwave}}
 Across all loss variants, we keep the network architecture and optimization settings fixed, following the original implementation. Specifically, we use:
\begin{itemize}
    \item \textbf{Training:}
    batch size $=16$,
    learning rate $=2{\times}10^{-4}$.

    \item \textbf{Data / preprocessing:}
    sample rate $=22050$ Hz,
    $n_\text{mels}=80$,
    $n_\text{fft}=1024$,
    hop $=256$ samples,
    crop mel frames $=62$.

    \item \textbf{Model:}
    30 residual layers,
    64 residual channels,
    dilation cycle length $=10$,
    conditional training.

    \item \textbf{Diffusion schedules:}
    inference noise schedule $=[10^{-4}, 10^{-3}, 10^{-2}, 0.05, 0.2, 0.5]$.
\end{itemize}

We fine-tune DiffWave for 150,000 steps on the LJSpeech-1.1 dataset \citep{ljspeech}, which contains 13,100 short utterances from a single speaker reading passages from seven non-fiction books. Following the original DiffWave training protocol, we excluded the LJ001* and LJ002* subsets from the training split, which we used for evaluation.

\begin{figure}[p]
\centering
\includegraphics[width=0.9\linewidth]{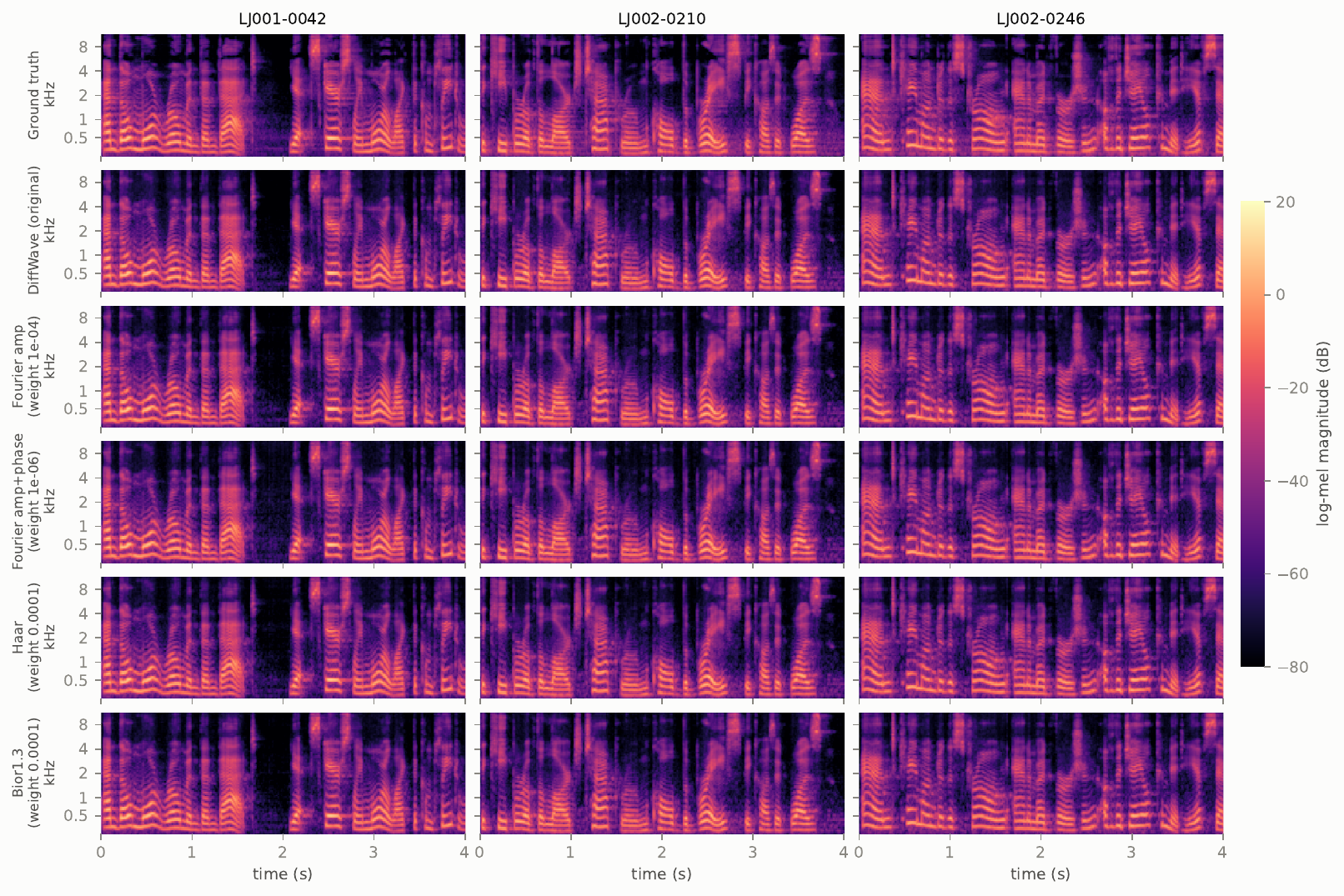}
\caption{Log-mel spectrograms (dB) of three held-out LJSpeech utterances. Rows show the ground-truth recording, the original DiffWave checkpoint, and the four spectrally regularized models, each at the weight $\lambda$ with the lowest FAD in Table~\ref{tab:audio_metrics}.}
\label{fig:diffwave_spectrograms}

\vspace{1.5em}

\includegraphics[width=0.9\linewidth]{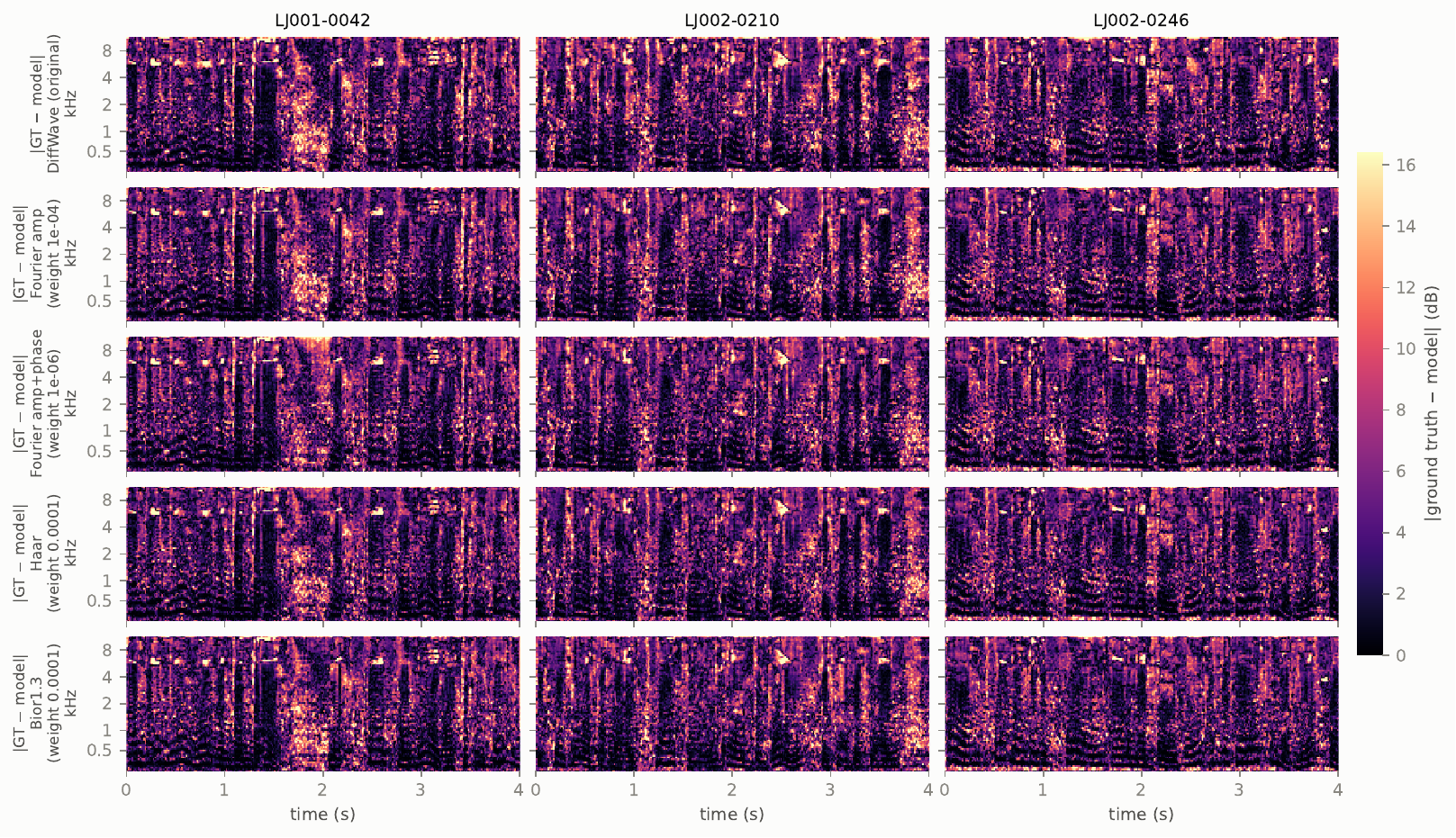}
\caption{Absolute log-mel error (dB) between the ground truth and each model for the utterances in Figure~\ref{fig:diffwave_spectrograms}. Brighter colors indicate larger error.}
\label{fig:diffwave_spectrogram_diff}
\end{figure}

\subsection{Generated Spectrograms}\label{app:diffwave_spectrograms}
 
To complement the aggregate metrics in Table~\ref{tab:audio_metrics}, we visualize log-mel spectrograms of waveforms generated by each fine-tuned model. Figure~\ref{fig:diffwave_spectrograms} shows three held-out LJSpeech utterances (LJ001-0042, LJ002-0210, and LJ002-0246), each synthesized from the same ground-truth mel conditioning by the original DiffWave checkpoint and by the four spectrally regularized models. For each regularizer we show the weight $\lambda$ that attains the lowest FAD in Table~\ref{tab:audio_metrics}, namely $\lambda = 10^{-4}$ for the Fourier amplitude, Haar, and Bior1.3 losses and $\lambda = 10^{-6}$ for the Fourier amplitude and phase loss.
 
All models reproduce the harmonic stacks and formant trajectories of the voiced segments, so the differences between methods are subtle at this scale. Figure~\ref{fig:diffwave_spectrogram_diff} therefore plots the absolute log-mel error $\lvert S_{\mathrm{GT}} - S_{\mathrm{model}} \rvert$ in dB for the same utterances. The error is small along the low-frequency harmonics and concentrates in low-energy regions, in particular the bands above roughly $4$~kHz and the pauses between words, where the logarithmic scale amplifies small absolute deviations. Because sampling is stochastic, part of this residual reflects sample-to-sample variability in unvoiced and silent regions rather than a systematic spectral bias.
 
\subsection{Power Spectra of the Fine-Tuned Models}\label{app:image_spectra}

\begin{figure}[t]
\centering
\includegraphics[width=0.85\linewidth]{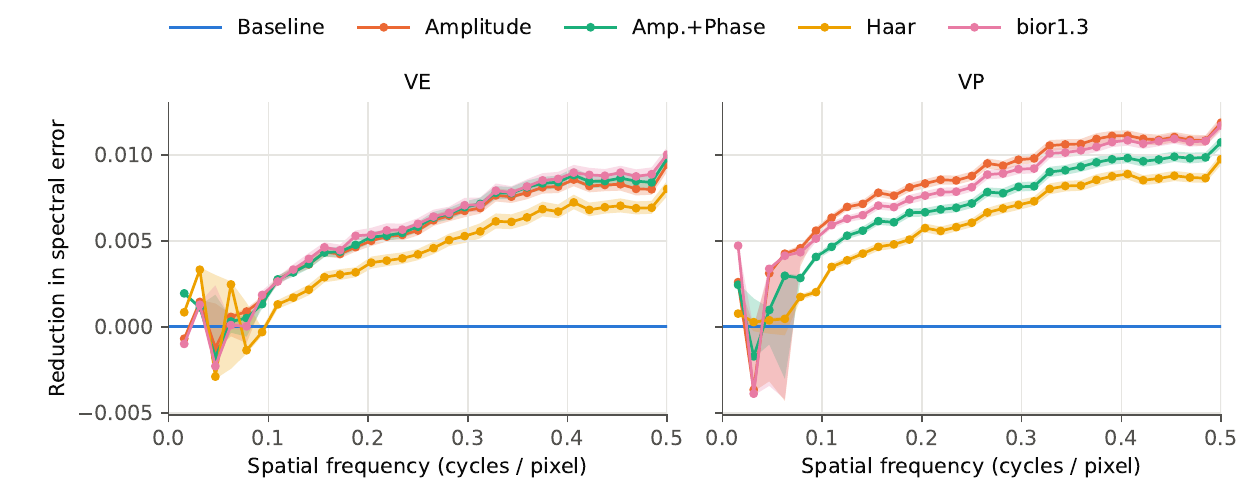}
\caption{Reduction in spectral error relative to the pretrained EDM baseline on FFHQ $64 \times 64$, for the VE (left) and VP (right) variants. Values above zero indicate a radially averaged power spectrum closer to that of the training images. All losses are shown in the unweighted setting, $\lambda = 1$. Each curve is computed on 10,000 samples drawn from the same seeds as the baseline, and the shaded bands are pointwise $95\%$ paired bootstrap intervals over these samples.}
\label{fig:spectra_ffhq}
\end{figure}

\begin{figure}[t]
\centering
\includegraphics[width=0.85\linewidth]{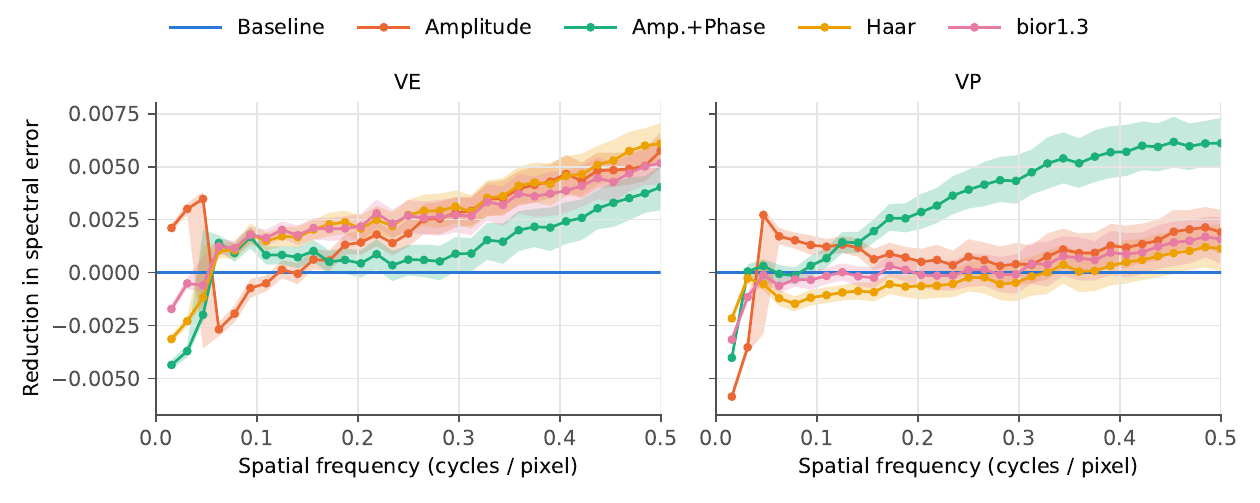}
\caption{Reduction in spectral error relative to the pretrained EDM baseline on AFHQv2 $64 \times 64$, for the VE (left) and VP (right) variants. Values above zero indicate a radially averaged power spectrum closer to that of the training images. All losses are shown in the unweighted setting, $\lambda = 1$. Each curve is computed on 10,000 samples drawn from the same seeds as the baseline, and the shaded bands are pointwise $95\%$ paired bootstrap intervals over these samples.}
\label{fig:spectra_afhq}
\end{figure}

\begin{figure}[t]
\centering
\includegraphics[width=0.85\linewidth]{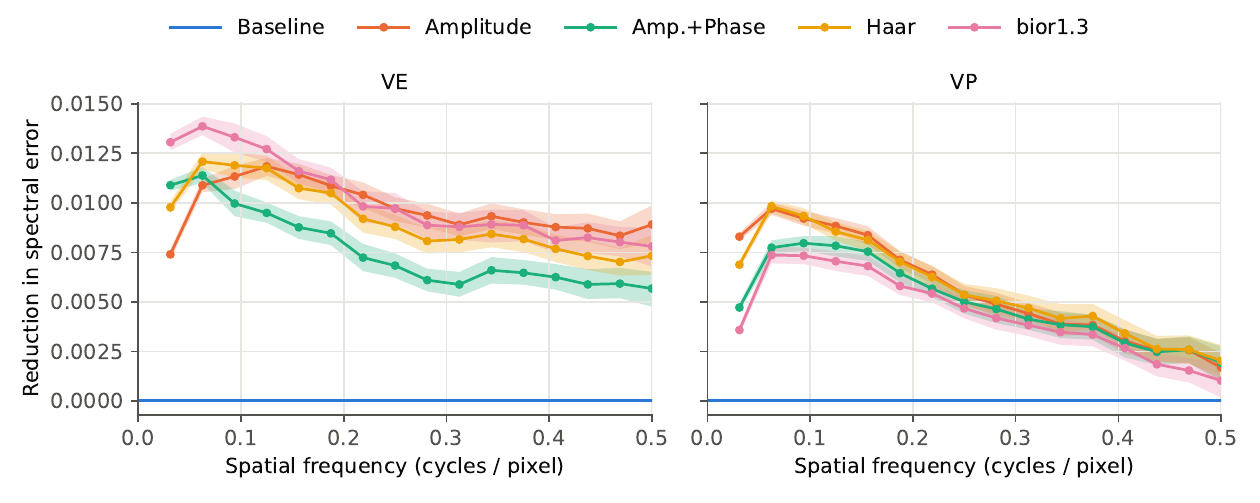}
\caption{Reduction in spectral error relative to the pretrained EDM baseline on class-conditional CIFAR-10 $32 \times 32$, for the VE (left) and VP (right) variants. Values above zero indicate a radially averaged power spectrum closer to that of the training images. All losses are shown in the unweighted setting, $\lambda = 1$. Each curve is computed on 10,000 samples drawn from the same seeds as the baseline, and the shaded bands are pointwise $95\%$ paired bootstrap intervals over these samples.}
\label{fig:spectra_cifar}
\end{figure}

To complement the FID scores in Table~\ref{tab:image_fid}, we compare the radially averaged power spectrum $P(r)$ of the generated images (Appendix~\ref{app:distill-eval}) with that of the training images. We define the spectral error of a model as $e(r) = \lvert P_{\mathrm{gen}}(r) / P_{\mathrm{real}}(r) - 1 \rvert$, where radius $r$ corresponds to $r/N$ cycles per pixel. Figures~\ref{fig:spectra_ffhq}--\ref{fig:spectra_cifar} plot the reduction $e_{\mathrm{EDM}}(r) - e_{\mathrm{reg}}(r)$ relative to the pretrained checkpoint, so that values above zero indicate a spectrum closer to the real data. Both models are sampled from the same 10,000 seeds, and the shaded bands are $95\%$ paired bootstrap intervals over these samples.

On FFHQ and AFHQ the pretrained models generate less power than the training images above $0.2$ cycles per pixel, by $7$--$8\%$ and about $19\%$, respectively. On FFHQ the models fine-tuned with a spectral loss have more power in this range for both EDM variants, and the reduction in error grows with frequency. Averaged over the radii up to $N/2$, the spectral error decreases by $0.004$--$0.008$, which is $8$--$13\%$ of the error of the pretrained model, and the Haar loss gives a smaller reduction than the other three. On AFHQ the reduction is smaller. For the VE variant it is $0.001$--$0.003$, and for the VP variant only the Fourier amplitude and phase loss differs from the baseline, by $0.003$. On CIFAR-10 the pretrained models generate $5$--$8\%$ more power than the training images at every frequency, and the fine-tuned models lower this difference by $0.004$--$0.010$ across the whole spectrum.

\clearpage
\newpage

\section{Few-Step Distillation Details}
\label{app:distill}
This appendix gives the details of the distillation experiments in
Section~\ref{sec:distillation}. For each run we distill a student, sample 50k
images from it, and compute FID and power spectra. Some generated images are shown in Appendix~\ref{app: gen_image_afhq_ffhq_distill}.

\subsection{Configuration}
\label{app:distill-config}

We use the SFD code of \citet{zhou2024sfd} with its CIFAR-10 settings on both datasets
(Table~\ref{tab:distill-config}). The SFD code lists these settings as usable for its
EDM models on CIFAR-10 and FFHQ. The schedule is polynomial with $\rho = 7$ on
$[\sigma_{\min}, \sigma_{\max}] = [0.006, 80]$. This gives four time points,
$\sigma = (80, 10.93, 0.67, 0.006)$. SFD sets $\sigma_{\min} = 0.006$, where EDM uses
$0.002$. The teacher is DPM-Solver++ \citep{lu2022dpmsolverpp} with $M = 3$ intermediate points per student step. That is 13 time points and 12 NFE (number of function evaluations) in total. For one student step from $\sigma_n$ to $\sigma_{n+1}$ the per-sample distillation loss is
\begin{equation}
  \mathcal{L}^{(n)}_{\mathrm{SFD}} = \big\| x^{\mathrm{S}}_{\sigma_{n+1}} - x^{\mathrm{T}}_{\sigma_{n+1}} \big\|_1 ,
\end{equation}
summed over all pixel values ($3 \times 32 \times 32$ on CIFAR-10 and
$3 \times 64 \times 64$ on FFHQ) and averaged over the batch. The analytical first
step is enabled, so the loss of step $n = 0$ is not back-propagated. Gradients come
only from the final two steps.

\begin{table}[ht]
\centering\small
\caption{SFD configuration used for every distillation run.}
\label{tab:distill-config}
\begin{tabular}{l>{\raggedright\arraybackslash}p{0.7\linewidth}}
\toprule
Data and teacher & CIFAR-10 $32\times32$, unconditional, EDM VP checkpoint \texttt{edm-cifar10-32x32-uncond-vp}. FFHQ $64\times64$, unconditional, \texttt{edm-ffhq-64x64-uncond-vp} \\
Student & initialized from the teacher, Euler steps with the analytical first step, \texttt{num\_steps}${}=4$ (3 steps, 2 NFE) \\
Time points & polynomial schedule, $\rho = 7$, $\sigma \in [0.006, 80]$, giving $(80, 10.93, 0.67, 0.006)$ \\
Teacher solver & DPM-Solver++, order 3, data prediction, lower order at the end, $M = 3$ (13 points, 12 NFE) \\
Optimization & Adam, learning rate $5\times10^{-5}$, $\beta = (0.9, 0.999)$, learning rate divided by 10 after half of training \\
Budget & 200k trajectories (200 kimg), batch 128, one RTX 6000 Ada GPU, about 36 min per CIFAR-10 run and 88 min per FFHQ run including evaluation \\
\bottomrule
\end{tabular}
\end{table}

\subsection{Relative Size of the Loss Terms}
\label{app:distill-calib}

We placed each grid of weights using the size of the spectral term relative to the
pixel term. On paired samples this ratio is
\begin{equation}
  \lambda_{\mathrm{eq}} = \frac{\mathbb{E}\, \left[ \| x^{\mathrm{S}} - x^{\mathrm{T}} \|_1\right]}{\mathbb{E}\, \left[ \mathcal{S}(x^{\mathrm{S}}, x^{\mathrm{T}})\right]},
\end{equation}
so at weight $\lambda$ the spectral term is $\lambda / \lambda_{\mathrm{eq}}$ times
the pixel term. We measured $\lambda_{\mathrm{eq}}$ on CIFAR-10 with 2,000 pairs of
final-state samples. Each pair has one sample from a trained $\lambda = 0$ student and
one from the teacher, drawn from the same noise. The ratio is also printed at the
first training step of every run (Table~\ref{tab:distill-calib}). Relative to the
pixel term, the bior1.3 term is 2.6 times the size of the amplitude term at the same
$\lambda$. Each grid covers about $10\%$, $33\%$ and $100\%$ of the pixel term. This
way no loss is compared only at weights where it dominates the objective. The
amplitude weights 1 and 3.5 come from a first sweep that predates this calibration. We
use the ratio only to choose the grids and report all results against $\lambda$. On
FFHQ the amplitude term at $\lambda = 1$ is about $23\%$ of the pixel term at the
first training step. On CIFAR-10 it is $29\%$. The carried-over weight has a similar
relative strength on both datasets.

\begin{table}[ht]
\centering\small
\caption{Size of each spectral term relative to the pixel term on CIFAR-10. ``At step
1'' is the mean $\pm$ SD over the first training step of every run using that loss.
``After training'' uses 2,000 paired samples and was used to place the grids.}
\label{tab:distill-calib}
\begin{tabular}{lcccc}
\toprule
 & \multicolumn{2}{c}{$\lambda_{\mathrm{eq}}$ (pixel over spectral term)} & & Spectral term as \\
\cmidrule(lr){2-3}
Loss & At step 1 & After training & $\lambda$ values run & share of pixel term (\%) \\
\midrule
Amplitude  & $3.50 \pm 0.07$ & 3.09 & 0.3, 1, 3.5 & 10, 32, 113 \\
Amp.+phase & $1.97 \pm 0.02$ & 2.06 & 0.2, 0.7, 2 & 10, 34, 97 \\
Haar       & $1.90 \pm 0.02$ & 1.29 & 0.13, 0.4, 1.3 & 10, 31, 101 \\
bior1.3    & $1.77 \pm 0.02$ & 1.19 & 0.12, 0.4, 1.2 & 10, 34, 101 \\
\bottomrule
\end{tabular}
\end{table}

\subsection{Evaluation and Statistics}
\label{app:distill-eval}

\paragraph{Runs.}
Every distillation setting, including the baseline, uses three independent training
seeds. The $k$-th run of each setting is paired with the baseline run using the same
training and sampling seeds. We report FID as the mean $\pm$ SD over the three runs,
and $\Delta$FID as the change in this mean from the baseline.

\paragraph{FID.}
We compute FID on 50,000 samples per run against the EDM training-set statistics. We
use the Inception-v3 network distributed with the StyleGAN3 metrics.

\paragraph{Power spectrum.}
For each run we compute the radially averaged power spectrum of its first 10,000
samples. For $N \times N$ images with $C$ channels,
\begin{equation}
  P(r) = \frac{1}{C\, |R_r|}\, \mathbb{E}_{x} \sum_{c=1}^{C} \sum_{k \in R_r} \big| \widetilde{X}_c(k) \big|^2,
  \qquad R_r = \big\{ k \in \{-\tfrac{N}{2}, \dots, \tfrac{N}{2} - 1\}^2 \,\big|\, \lfloor \|k\|_2 \rfloor = r \big\},
\end{equation}
for $r = 0, \dots, N/2$, where $\widetilde{X}_c$ is the centered, orthonormally scaled
two-dimensional DFT of channel $c$. Radius $r$ corresponds to $r/N$ cycles per pixel.
The teacher spectrum uses 10,000 samples from the same initial noise, generated
with the solver used during training. The plots show $\log_{10}$ power ratios
relative to the mean baseline spectrum. Each loss curve summarizes the three runs
of its setting, and the shaded band shows the range of the three baseline spectra
relative to their mean. Zero denotes the baseline mean, positive values indicate
more power, and negative values indicate less power. The teacher curve provides a
reference at each frequency. Moving toward that curve indicates closer agreement
in average power at that frequency, but does not establish better individual
samples. Frequencies are expressed in cycles per pixel for both datasets.

\subsection{CIFAR-10 Results}
\label{app:distill-cifar}

\paragraph{Baseline.}
Unmodified SFD reproduces the published result. \citet{zhou2024sfd} report FID 4.53.
The three baseline runs used below give $4.475 \pm 0.020$.

\paragraph{All settings.}
Table~\ref{tab:distill} gives FID for every setting. The amplitude loss lowers FID at $\lambda = 0.3$ and $\lambda = 1$ and raises it at $\lambda = 3.5$. Both wavelet losses raise FID at every weight, and more at larger weights. This holds even at their smallest weights, where the wavelet term is only about $10\%$ of the pixel term. Haar and bior1.3 give nearly the same results. Amplitude with phase gives a small improvement at $\lambda = 0.2$ and hurts at $0.7$ and $2$. It is worse than amplitude alone at comparable weights.

\begin{figure}[!htbp]
\centering
\begin{subfigure}[t]{0.49\linewidth}
\centering
\includegraphics[width=\linewidth]{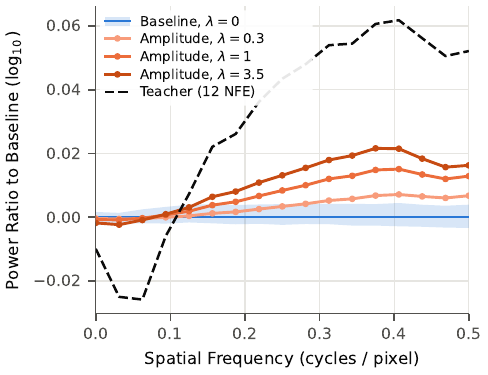}
\caption{Amplitude}
\label{fig:distill-rapsd-amp}
\end{subfigure}\hfill
\begin{subfigure}[t]{0.49\linewidth}
\centering
\includegraphics[width=\linewidth]{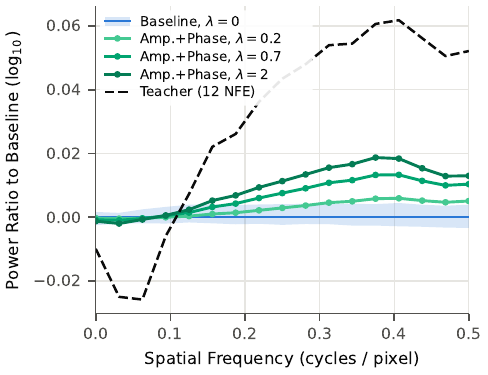}
\caption{Amplitude with phase}
\label{fig:distill-rapsd-ampphase}
\end{subfigure}

\medskip
\begin{subfigure}[t]{0.49\linewidth}
\centering
\includegraphics[width=\linewidth]{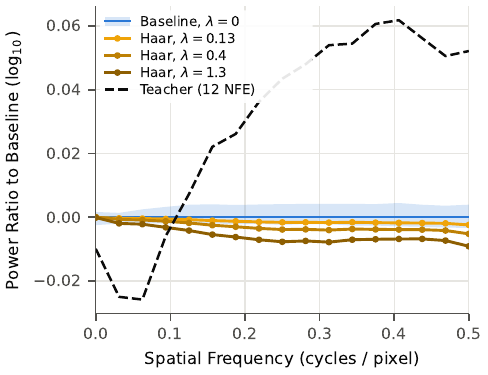}
\caption{Haar}
\label{fig:distill-rapsd-haar}
\end{subfigure}\hfill
\begin{subfigure}[t]{0.49\linewidth}
\centering
\includegraphics[width=\linewidth]{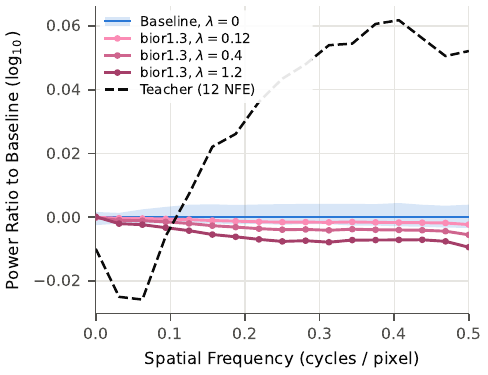}
\caption{bior1.3}
\label{fig:distill-rapsd-bior13}
\end{subfigure}
\caption{Power spectra of 2-NFE CIFAR-10 students for each loss at each of its weights (darker is larger) and of the 12-NFE teacher. Curves show the $\log_{10}$ power ratio to the mean baseline spectrum. The shaded band shows the range of the three baseline spectra relative to their mean. Each loss curve summarizes three runs. The Fourier
losses raise mid- and high-frequency power toward the teacher's, more at larger
weights. The wavelet losses lower it. Haar and bior1.3 are close at matching weights.}
\label{fig:distill-rapsd-sweeps}
\end{figure}

\begin{figure}[!htbp]
\centering
\begin{subfigure}[t]{0.49\linewidth}
\centering
\includegraphics[width=\linewidth]{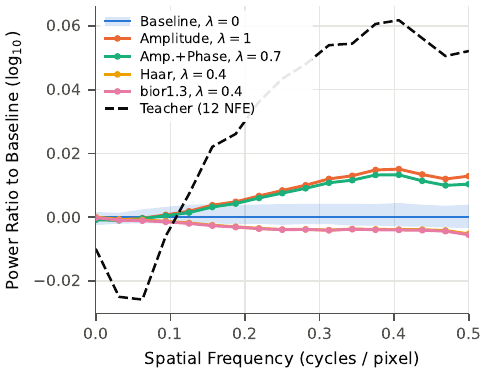}
\caption{CIFAR-10}
\label{fig:distill-rapsd-types}
\end{subfigure}\hfill
\begin{subfigure}[t]{0.49\linewidth}
\centering
\includegraphics[width=\linewidth]{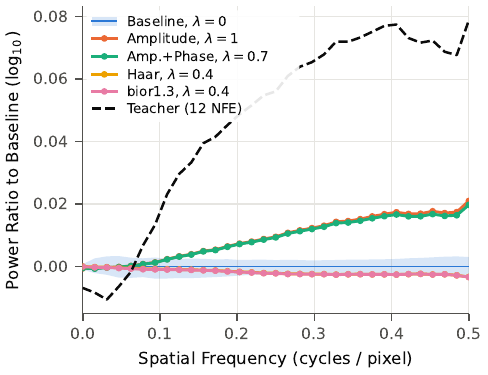}
\caption{FFHQ $64\times64$}
\label{fig:distill-rapsd-ffhq}
\end{subfigure}
\caption{The four losses at their middle weights (amplitude $\lambda = 1$, amplitude
with phase $0.7$, Haar and bior1.3 $0.4$), shown as in
Figure~\ref{fig:distill-rapsd-sweeps}.
Fourier losses raise high-frequency power toward the teacher's and the wavelet losses lower it. On FFHQ the two Fourier curves nearly overlap, as do the two wavelet curves.}
\label{fig:distill-rapsd-mid}
\end{figure}

\paragraph{Spectra.}
Figures~\ref{fig:distill-rapsd-sweeps} and~\ref{fig:distill-rapsd-mid} compare the
regularized students with the baseline and teacher. On CIFAR-10, the teacher has
more mid- and high-frequency power than the baseline student and less low-frequency
power. The Fourier losses raise mid- and high-frequency power toward the teacher's,
with larger changes at larger weights. The wavelet losses lower power in these
bands and increase the deficit relative to the teacher. Haar and bior1.3 produce
similar spectral changes at corresponding weights.

For amplitude at $\lambda=0.3$, mean FID decreases by $0.028$, which is smaller than
the run-to-run standard deviation at that setting ($0.038$). Increasing the amplitude
weight from $\lambda=1$ to $3.5$ brings high-frequency power closer to the teacher's,
while FID worsens. Matching the teacher's average power spectrum therefore does not
necessarily improve sample quality as measured by FID.

\subsection{FFHQ \texorpdfstring{$64\times64$}{64x64}}
\label{app:distill-ffhq}

We distilled the EDM FFHQ teacher with every setting in Table~\ref{tab:distill-config}
unchanged. FID is computed against the EDM FFHQ statistics. We ran the baseline and
the amplitude loss at $\lambda = 1$ first. Nothing was selected on FFHQ, so this tests
a setting and a direction of effect fixed in advance. The other three losses came
next, each at the middle weight of its CIFAR-10 grid (amplitude with phase $0.7$, Haar
and bior1.3 $0.4$). At these weights each term has about the same size relative to the
pixel term as amplitude at $\lambda = 1$. These weights were also fixed before the
runs. Finally we ran each loss three times at each of the other two weights of its
CIFAR-10 grid (24 runs). SFD does not report FFHQ results, so the baseline is our own
run of its code. Training takes 23.8\,s per kimg at batch 128 and uses at most 22\,GB
of GPU memory.

\paragraph{Amplitude.}
Across three runs, mean FID decreases by $0.104$ (approximately $1.8\%$). The corresponding decrease on CIFAR-10 is $0.052$ (approximately $1.2\%$). The pixel loss is unchanged at the reported precision ($0.1155$).

\paragraph{The other losses.}
Amplitude with phase at $\lambda=0.7$ lowers FID by $0.107$, similar to the $0.104$ decrease from amplitude alone at $\lambda=1$. On CIFAR-10, amplitude with phase at the same weight instead increases FID by $0.014$.
Haar and bior1.3 at $\lambda=0.4$ increase FID by $0.082$ and $0.081$, respectively. Both increases are approximately $1.4\%$, compared with $4.3\%$ and $4.5\%$ on CIFAR-10. The two wavelets again give nearly the same FID.

\paragraph{Weights.}
Table~\ref{tab:distill-ffhq-grid} reports FID on FFHQ using the CIFAR-10 weight
grids, with three runs per setting. Both Fourier losses lower FID at every weight.
Both wavelet losses raise it, more at larger weights but less than on CIFAR-10.

The weight at which FID turns is different. From the middle to the largest weight, FID
rises on CIFAR-10 by $+0.077$ for amplitude and by $+0.137$ for amplitude with phase. On FFHQ, amplitude FID decreases by $0.021$, while amplitude with phase remains essentially unchanged. Amplitude has its lowest FFHQ FID at $\lambda = 3.5$, but the difference to $\lambda = 1$ is small. Choosing $\lambda = 3.5$ would also select the weight on the runs that evaluate it. We keep the carried-over $\lambda = 1$ as the main result.

At their middle weights, the losses shift the average spectra in similar directions
on both datasets (Figure~\ref{fig:distill-rapsd-mid}), while their effects on FID
depend on the dataset. The weight should be chosen by sample quality on each dataset.

\begin{table}[ht]
\centering\small
\caption{FID on FFHQ $64\times64$ at the CIFAR-10 weights. Values are mean
$\pm$ SD over three runs per setting, with 50,000 generated images per run.}
\label{tab:distill-ffhq-grid}
\begin{tabular}{lrc}
\toprule
Loss & $\lambda$ & FID $\downarrow$ \\
\midrule
Baseline & 0 & $5.837 \pm 0.016$ \\
\midrule
Amplitude & 0.3 & $5.777 \pm 0.032$ \\
Amplitude & 1 & $5.733 \pm 0.051$ \\
Amplitude & 3.5 & $5.712 \pm 0.053$ \\
\midrule
Amp.+phase & 0.2 & $5.781 \pm 0.032$ \\
Amp.+phase & 0.7 & $5.730 \pm 0.050$ \\
Amp.+phase & 2 & $5.731 \pm 0.042$ \\
\midrule
Haar & 0.13 & $5.869 \pm 0.039$ \\
Haar & 0.4 & $5.919 \pm 0.061$ \\
Haar & 1.3 & $6.001 \pm 0.097$ \\
\midrule
bior1.3 & 0.12 & $5.864 \pm 0.034$ \\
bior1.3 & 0.4 & $5.918 \pm 0.056$ \\
bior1.3 & 1.2 & $5.986 \pm 0.091$ \\
\bottomrule
\end{tabular}
\end{table}

\paragraph{Spectra.}
Figure~\ref{fig:distill-rapsd-ffhq} shows the FFHQ power ratios at the middle
weights. As on CIFAR-10, the baseline student has less mid- and high-frequency
power than the teacher. The Fourier losses raise power in these bands toward the
teacher's, while the wavelet losses lower it further. The two Fourier curves nearly
overlap, as do the two wavelet curves. The teacher and baseline differ less at low
frequencies on FFHQ than on CIFAR-10.

\subsection{Discussion}
\label{app:distill-discussion}

The wavelet result differs from fine-tuning, where all four losses gave similar FID
(see Appendix~\ref{sec:image-exps}). One explanation is that a coefficient-wise $L^1$ penalty pulls a coefficient the student cannot predict toward the median of its possible values. We have not tested this directly. Pixels of fine texture and wavelet details are signed, localized coefficients. Their median is near zero when the student cannot resolve where the detail sits. This would explain the high-frequency deficit of the pixel-$L^1$ baseline and the further loss from a wavelet term. The Fourier amplitude is invariant to translation, so its target stays well defined under positional uncertainty. It ignores phase, however, and can supply energy with the wrong structure, as at large $\lambda$.

The relative gain on FFHQ at 2 NFE ($-1.8\%$) is smaller than the fine-tuning gain at 40 sampling steps for the same model and loss (Table~\ref{tab:image_fid}, $-2.9\%$). The two are not directly comparable. The fine-tuning baseline is the released checkpoint rather than a model fine-tuned with $\lambda = 0$. Still, these results do not show that spectral regularization helps more with fewer sampling steps.

\subsection{Limitations}
\label{app:distill-limits}

All experiments use one distillation method. The wavelet effects are large next to their run-to-run spread. The smaller amplitude-phase effects on CIFAR-10 are comparable to it. On FFHQ, amplitude achieves its lowest mean FID at the largest weight tested, so the optimum may lie outside the evaluated range. The spectral reference is the teacher, which is itself a 12-NFE sampler. We have not measured where the training images sit on these spectral axes.

FID is the only sample-quality measure. The regularizer changes the high-frequency
statistics to which Inception features are sensitive. A second feature space, such as Fr\'echet distance in DINOv2 features, together with precision and recall would guard against a metric artifact. The CIFAR-10 results show that bringing high-frequency power closer to the teacher's does not necessarily improve FID.

\clearpage
\newpage

\section{Graph Experimental Details}
\label{app:graph}
%
This appendix gives the details of the graph experiments in
Section~\ref{sec:graph}. We fine-tuned nine models, three weights with three
training seeds each, and evaluated each of them at five sampling budgets with
three sampling seeds. Every evaluation generates 10{,}000 molecules.

\subsection{Training Configuration}
\label{app:graph-setup}

We use the SwinGNN code of \citet{yan2024swingnn} and its released QM9
checkpoint \texttt{qm9\_scalar\_dim\_60}, which their paper calls SwinGNN
(scalar) (Table~\ref{tab:graph-config}). SwinGNN is an EDM model that denoises
the adjacency matrix and the atom types of a molecule together. We fine-tune it
from its released weights. The released batch of 15{,}360 molecules does not fit
on one 48~GB GPU, so we use a batch of 5{,}120.

\begin{table}[!htbp]
\centering\small
\caption{Configuration used for every fine-tune and evaluation.}
\label{tab:graph-config}
\begin{tabular}{l>{\raggedright\arraybackslash}p{0.75\linewidth}}
\toprule
Checkpoint & \texttt{qm9\_scalar\_dim\_60}: SwinGNN with feature dimension 60 and 15.3M parameters, released after 5{,}001 epochs at batch 15{,}360 \\
Data & QM9 \citep{ramakrishnan2014qm9} preprocessed as in GDSS \citep{jo2022score}: 133{,}885 kekulized molecules with 1 to 9 heavy atoms (C, N, O, F) and implicit hydrogens, of which 120{,}803 are for training and 13{,}082 form the test set \\
Encoding & bond order $k \in \{0, 1, 2, 3\}$ stored as $2k/3 - 1$, atom type as a scalar channel, molecules padded to 16 atoms \\
Fine-tuning & from the released EMA weights; Adam, constant learning rate $5 \times 10^{-5}$, batch 5{,}120, 500 epochs ($\approx 12{,}000$ steps), gradient clipping at norm 10 \\
Spectral term & eigenvalue loss on the predicted bond orders at every noise level, $\lambda = 0.0176$, $0.0586$ and $0.1757$, three training seeds each \\
Sampler & EDM stochastic Heun, $\sigma \in [0.002, 80]$, $\rho = 7$, $S_{\mathrm{churn}} = 40$; $N$ steps cost $2N - 1$ NFE \\
Evaluation & EMA 0.999 weights, 10{,}000 molecules per evaluation\\
Budget & one RTX 6000 Ada GPU; about 2.9 hours per fine-tune, and 2.4 to 3.8 minutes per evaluation at $N \le 64$ and 8 minutes at $N = 256$ \\
\bottomrule
\end{tabular}
\end{table}

\paragraph{EMA rate.}\label{app:graph-ema} SwinGNN keeps an exponential moving
average (EMA) of the weights and evaluates the released model with rate 0.9999.
We therefore evaluate the fine-tunes with rate 0.999, which we chose before evaluating any of them, and evaluate the released model with the same rate. The rate matters for the comparison. With rate 0.999 instead of 0.9999 the released model makes fewer invalid molecules, $76.7$, $46.0$, $20.7$ and $10.7$ per $10^4$ with 16, 32, 64 and 256 steps against $113.3$, $78.3$, $50.3$ and $27.0$, while its FCD changes by at most $0.011$. Comparing a fine-tune at rate 0.999 with the released model at 0.9999 would therefore credit the fine-tune with a validity gain that comes from the EMA.

\paragraph{Baseline.} Evaluated with the authors' code at 256 steps and its
default EMA rate, the released checkpoint gives $99.73\%$ validity without
correction and $96.28\%$ uniqueness, against the $99.68\%$ and $95.92\%$ reported
by \citet{yan2024swingnn} for SwinGNN. Its FCD and NSPDK are lower than
reported, $0.105$ against $0.169$ and $2.43 \times 10^{-4}$ against
$4.02 \times 10^{-4}$. We could not tell whether the checkpoint or newer package
versions cause this, so we compare against our own evaluation. As a reference,
10{,}000 molecules drawn from the training set score an FCD of $0.077 \pm 0.006$
against the test set.

\subsection{The Spectral Term and Its Weight}
\label{app:graph-loss}

The denoiser predicts the adjacency on the scale of the encoding. We map its
output $D_\theta$ back to bond orders,
\begin{equation}
  \widehat{A}_0 = \tfrac{3}{2}\big(\mathrm{clip}(D_\theta, -1, 1) + 1\big),
\end{equation}
make it symmetric and keep only the rows and columns of real atoms. On clean data
this recovers the bond orders exactly, so the diagonal of the Laplacian holds the
valence of each atom. Padded atoms add the same zero eigenvalues to both spectra,
so they do not affect the loss. We compute the eigenvalues of the $16 \times 16$
Laplacians with \texttt{torch.linalg.eigvalsh}, which is cheap, and add the
eigenvalue loss to the adjacency loss of SwinGNN at every noise level.

\paragraph{Weight.} We set $\lambda$ by the size of the spectral term relative to
the adjacency loss. Before training we measure both losses on 20 training
batches (102{,}400 molecules) at the starting weights and set
$\lambda = r \cdot \mathrm{mean}(\ell_{\mathrm{adj}}) / \mathrm{mean}(\ell_{\mathrm{G}})$.
With $r = 0.03$, $0.1$ and $0.3$, so that the spectral term starts at $3\%$,
$10\%$ and $30\%$ of the adjacency loss, this gave us values of $\lambda = 0.0176$, $0.0586$ and $0.1757$.

\subsection{Evaluation}
\label{app:graph-eval}

\paragraph{Metrics.} We use the evaluation code of SwinGNN, which follows GDSS.
A molecule is invalid if RDKit cannot sanitize it as generated. Invalid molecules
are then repaired with the valence correction of GDSS, and the largest fragment
of each molecule is kept. On the repaired molecules we compute uniqueness,
novelty against the training set, FCD \citep{preuer2018fcd} and NSPDK MMD with
atom and bond labels \citep{costa2010nspdk}.

\paragraph{Runs and statistics.} Each trained model is sampled with three
sampling seeds, so each fine-tune cell has nine evaluations and each cell of the
released model three. With a given sampling seed, every model starts from the
same noise and generates molecules with the same numbers of atoms.
Table~\ref{tab:graph} gives the mean $\pm$ SD over the evaluations of a cell.

\paragraph{Held-out reference.} The noise weighting was chosen on one half of
the test set. Against the other half, which no choice used, FCD with 32 steps is
$0.335$ for the released model and $0.334$, $0.313$ and $0.311$ for the three
weights, and with 256 steps it is $0.125$ against $0.128$, $0.130$ and $0.167$.
These values are higher than in Table~\ref{tab:graph} because the reference is
smaller. The two larger weights are nearly tied at 32 steps, and the conclusions
are otherwise unchanged.

\subsection{Generated Molecules}
\label{app:graph-samples}

\begin{figure}[p]
\centering
\includegraphics[width=0.74\linewidth]{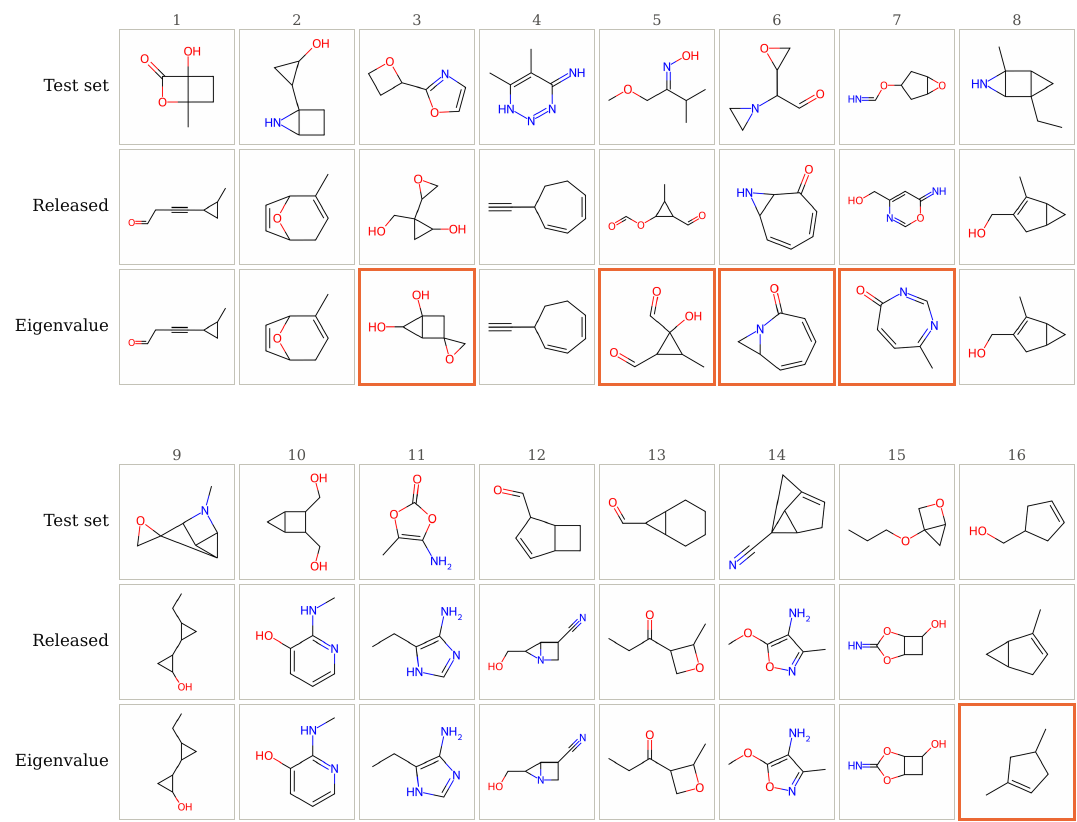}\\[2pt]
{\small (a) 256 steps}\\[8pt]
\includegraphics[width=0.74\linewidth]{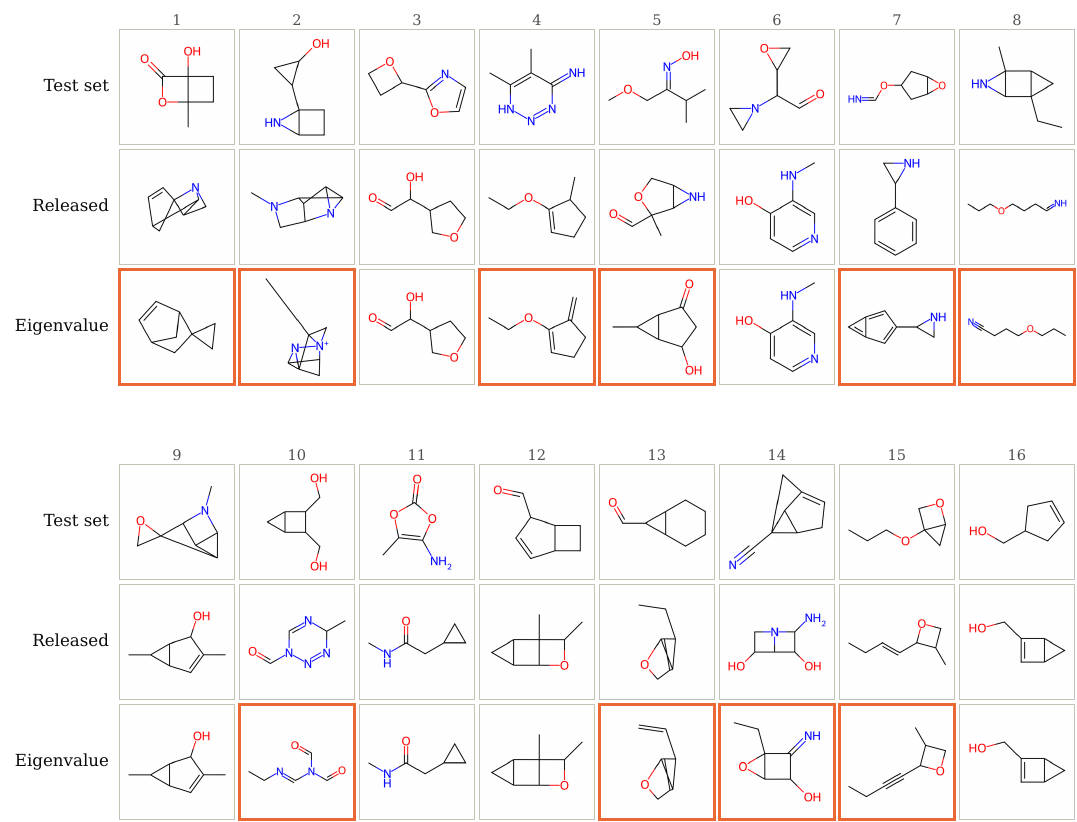}\\[2pt]
{\small (b) 16 steps}
\caption{Uncurated QM9 samples with (a) 256 and (b) 16 sampling steps. Each panel
shows the first 16 molecules of sampling seed 0 from the released model and from
its fine-tune with the eigenvalue loss ($r = 0.1$), both with EMA 0.999. The
molecules in a column start from the same noise and have as many atoms as the
test molecule at the top. An orange frame marks a molecule of the fine-tuned
model that differs from the released model's.}
\label{fig:graph-mol}
\end{figure}

Figures~\ref{fig:graph-mol} and~\ref{fig:graph-mol} show the first 16
molecules of sampling seed 0 from the released model and from the fine-tune with
$r = 0.1$, with 256 and 16 steps, without selection. None of them needed the
valence correction. Both models start from the same noise and generate the same
number of atoms in each column, taken from the test molecule at the top, so the
molecules in a column differ only because of the fine-tuning. Over all 10{,}000
samples the fine-tune produces the released model's molecule in $31.2\%$ of cases
with 256 steps and $28.9\%$ with 16 steps. Sixteen samples cannot show the
differences in Table~\ref{tab:graph}, which concern a few percent of the molecules.

\subsection{Limitations}
\label{app:graph-scope}

We have only used one dataset, one released checkpoint and one spectral loss, at three different weights. The released model has a single set of weights, so comparisons with it rest on three sampling seeds. Our FCD and NSPDK for the
released model differ from those reported by \citet{yan2024swingnn}, so all
comparisons are internal to our evaluation.
\clearpage
\newpage

\section{More Wavelet Transform Background} \label{app:wavelets}
The discrete wavelet transform represents a signal through localized approximation and detail coefficients at multiple scales. In one dimension, a multilevel decomposition produces a coarse approximation $a_J$ and detail coefficients $d_J,\ldots,d_1$. For images, a separable two-dimensional transform produces one approximation sub-band and three directional detail sub-bands at each decomposition level.

Let $\mathcal{W}$ denote the finite-dimensional wavelet analysis operator, including its boundary treatment and coefficient selection. The regularizer compares corresponding coefficients of the prediction $\hat{x}_0$ and target $x_0$,
\begin{equation}
\ell_{\mathrm{W}}(\hat{x}_0,x_0)
=
\sum_{s,\ell}\gamma_{s,\ell}
\left\|
(\mathcal{W}\hat{x}_0)^{(s,\ell)}
-
(\mathcal{W}x_0)^{(s,\ell)}
\right\|_1,
\end{equation}
where $s$ indexes scale, $\ell$ indexes subband, and the nonnegative weights $\gamma_{s,\ell}$ are fixed. The training regularizer is the expectation of this discrepancy over the training inputs.

The Haar transform computes local averages and differences using short filters. With compatible normalization and boundary treatment, it is orthonormal and preserves Euclidean norm. The biorthogonal 1.3 (bior1.3) transform uses distinct analysis and synthesis filters that support perfect reconstruction, but its analysis operator is not orthonormal. Accordingly, our analysis uses the operator bound
\begin{equation}
\|\mathcal{W}v\|_2 \le C_W\|v\|_2.
\end{equation}
An orthonormal Haar implementation has $C_W=1$, while the constant for a biorthogonal implementation depends on the finite-dimensional transform.

Wavelet discrepancies retain spatial or temporal localization and coefficient signs. Fourier amplitude discrepancies instead compare global spectral magnitudes and are invariant to circular translations. These properties define different training objectives, whose effects on generative quality are evaluated empirically.
\newpage

\section{Generated Images for AFHQ and FFHQ from EDM Fine-Tuning} \label{app: gen_image_afhq_ffhq}
Here we present some selected image samples from our models. 
\begin{figure}[t]
  \centering
  \begin{subfigure}{\columnwidth}
    \centering
    \includegraphics[scale = 0.4]{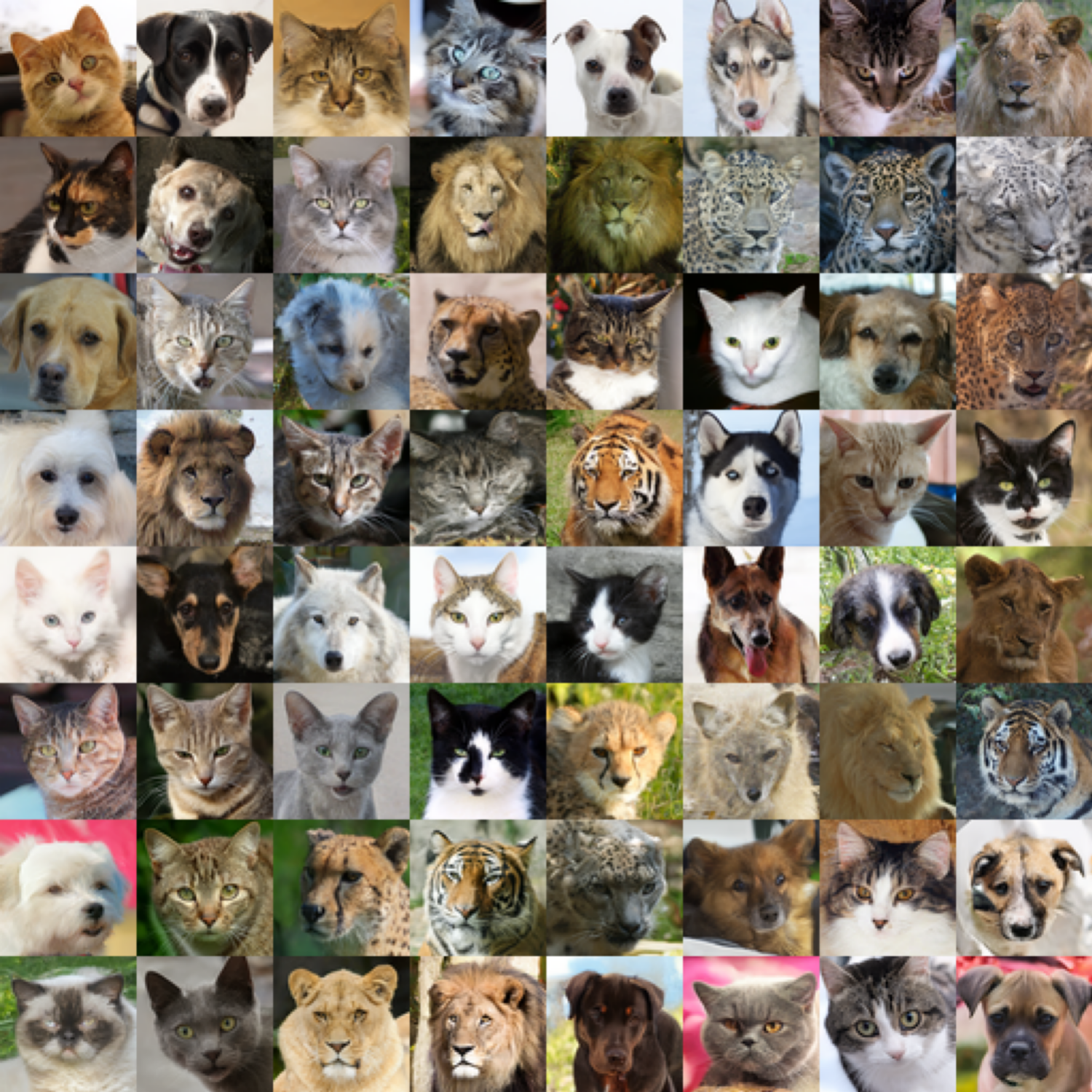}
    \caption{VE-EDM results}
  \end{subfigure}
  \vspace{0.8em}
  \begin{subfigure}{\columnwidth}
    \centering
    \includegraphics[scale = 0.4]{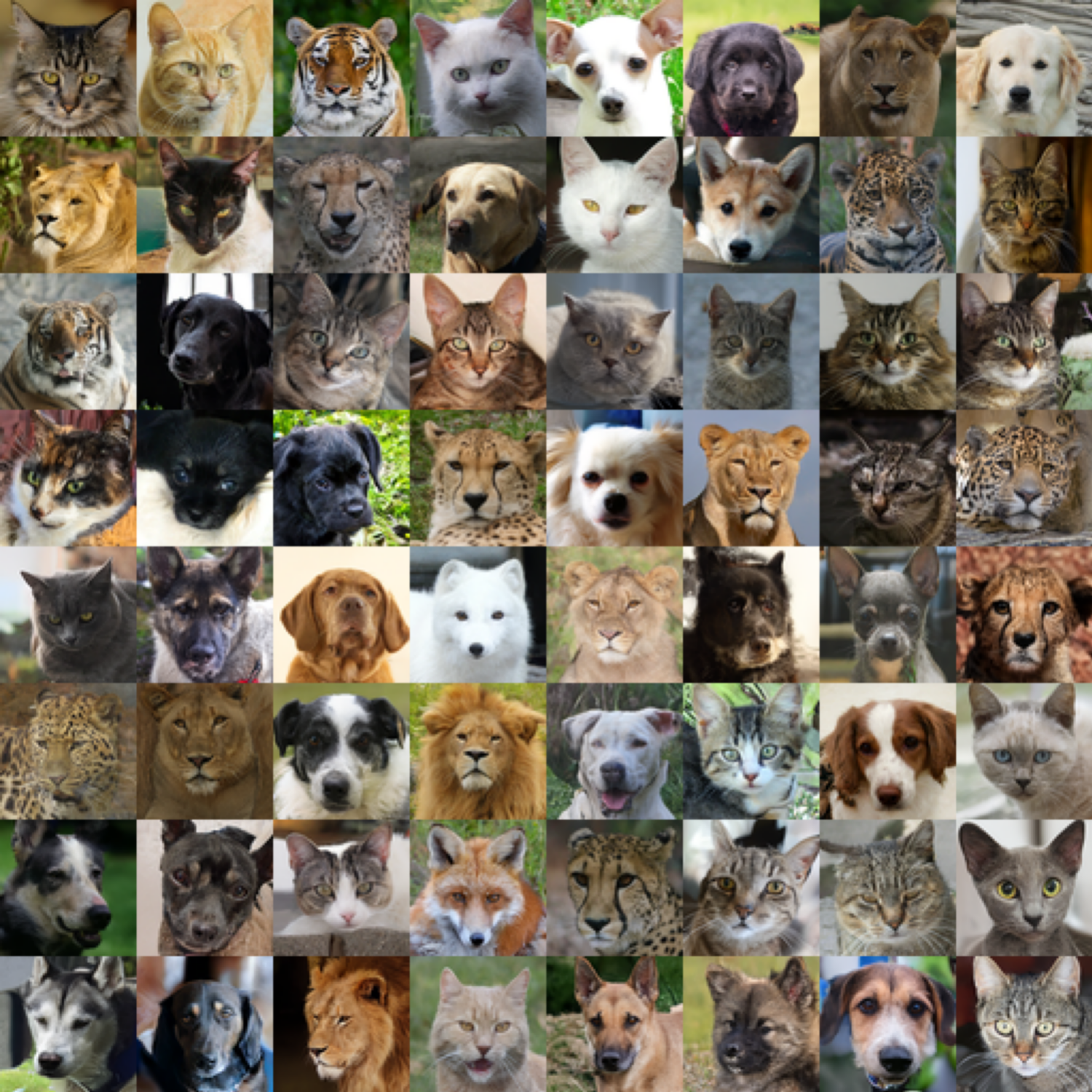}
    \caption{VP-EDM results}
  \end{subfigure}
  \caption{Generated AFHQ samples obtained by fine-tuning with the unweighted Fourier amplitude loss under different EDM formulations.}
  \label{fig:afhq_amp}
\end{figure}

\begin{figure}[t]
  \centering
  \begin{subfigure}{\columnwidth}
    \centering
    \includegraphics[scale = 0.4]{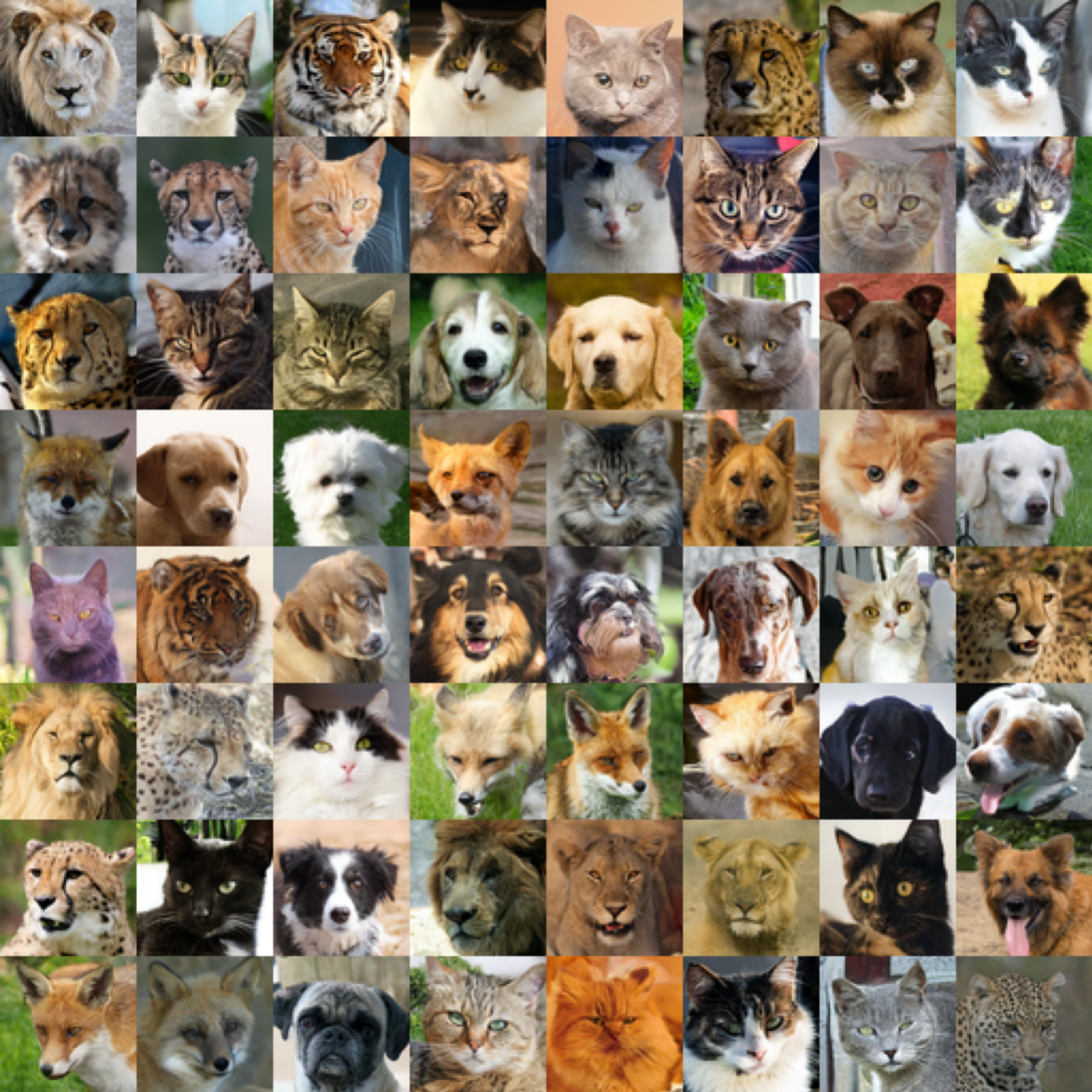}
    \caption{VE-EDM results}
  \end{subfigure}
  \vspace{0.8em}
  \begin{subfigure}{\columnwidth}
    \centering
    \includegraphics[scale = 0.4]{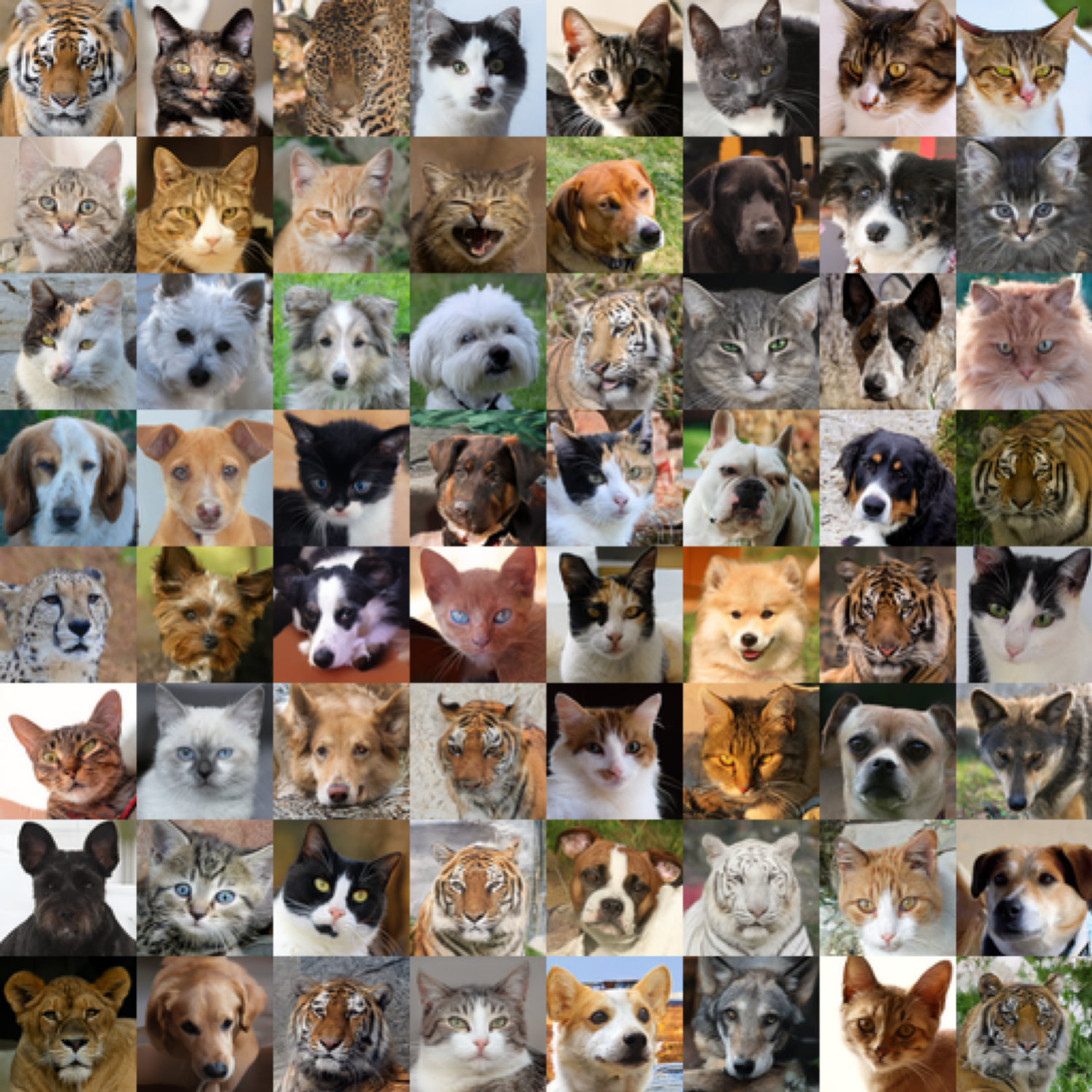}
    \caption{VP-EDM results}
  \end{subfigure}
  \caption{Generated AFHQ samples obtained by fine-tuning with the unweighted Fourier amplitude+phase loss under different EDM formulations.}
  \label{fig:afhq_amp_phase}
\end{figure}
\begin{figure}[t]
  \centering
  \begin{subfigure}{\columnwidth}
    \centering
    \includegraphics[scale = 0.4]{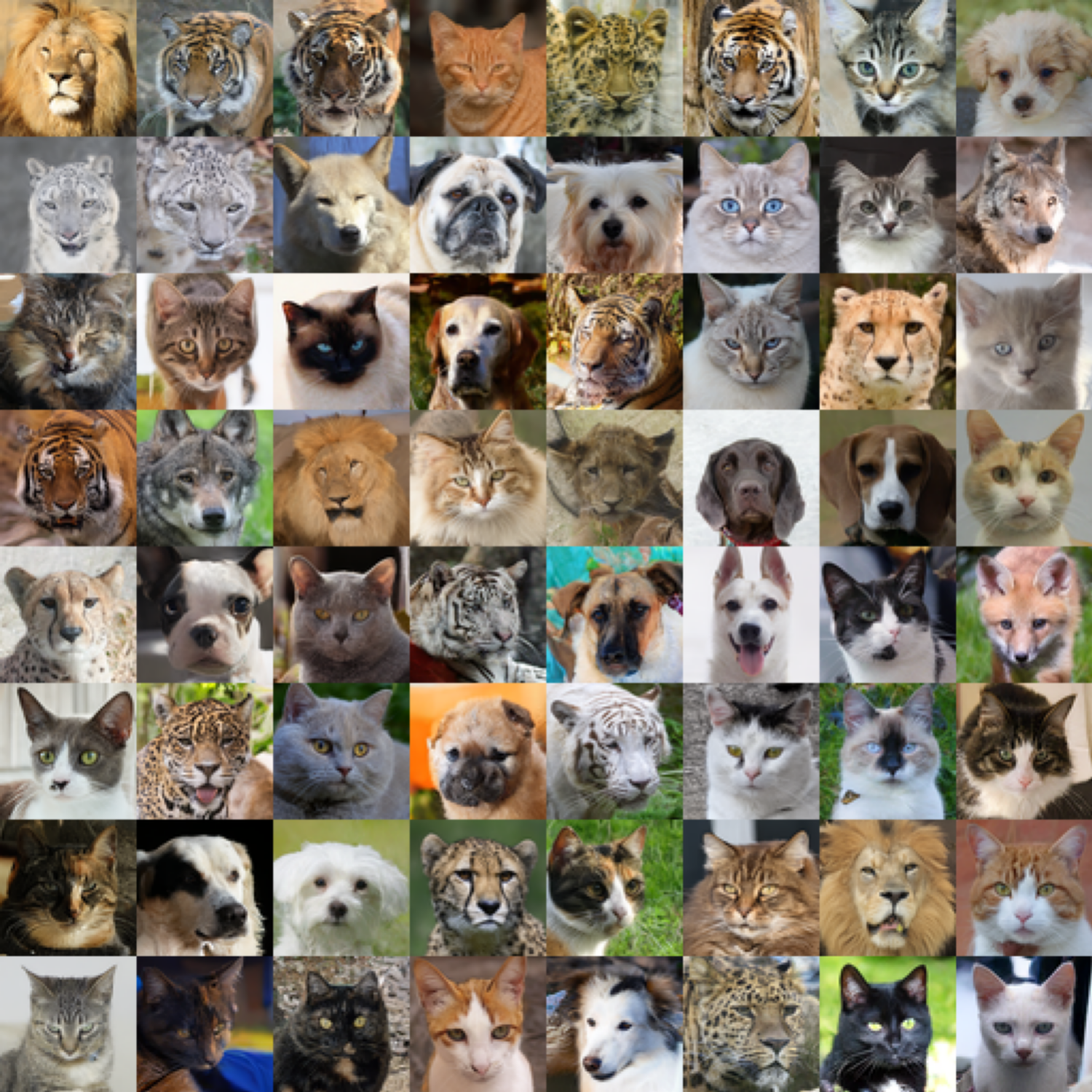}
    \caption{VE-EDM results}
  \end{subfigure}
  \vspace{0.8em}
  \begin{subfigure}{\columnwidth}
    \centering
    \includegraphics[scale = 0.4]{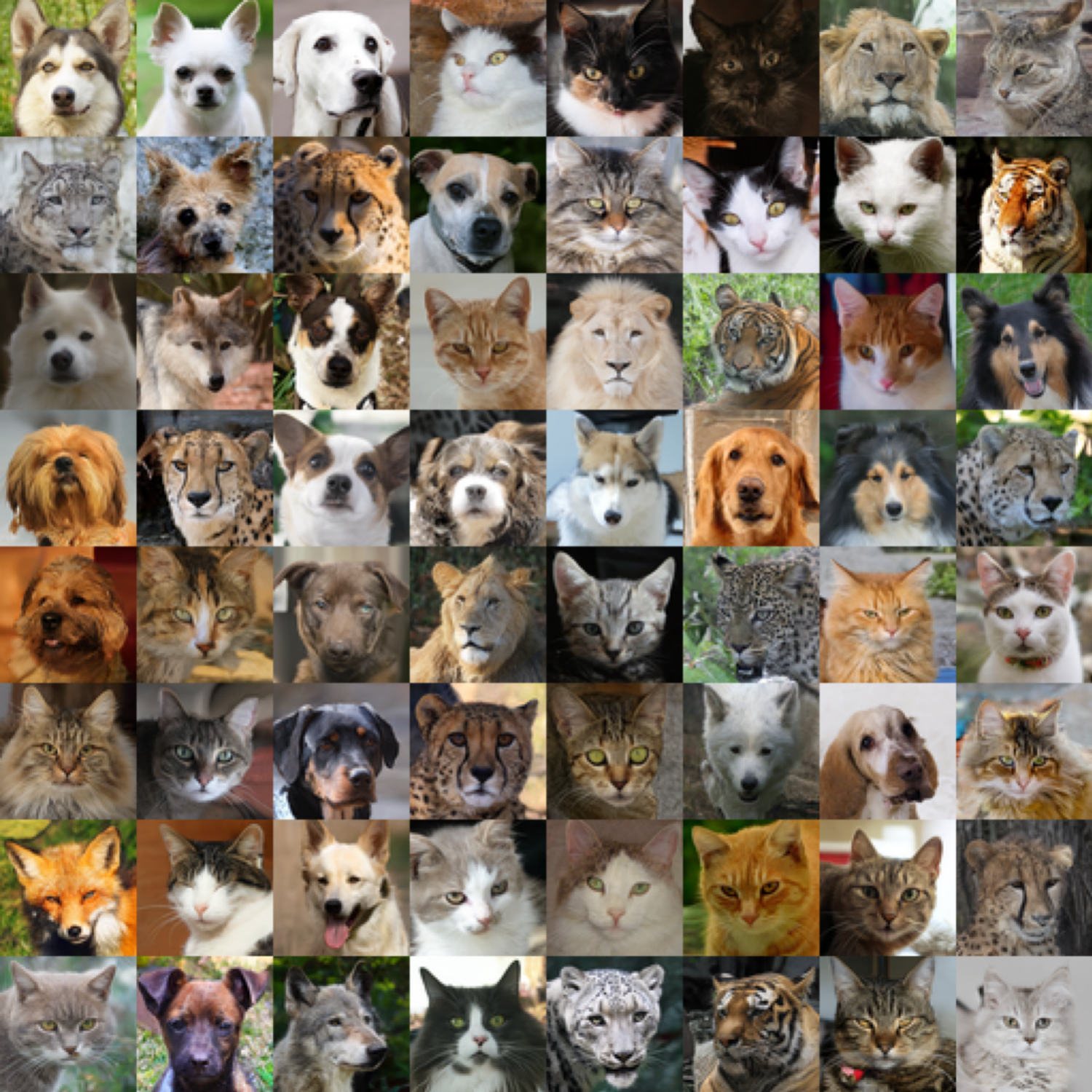}
    \caption{VP-EDM results}
  \end{subfigure}
  \caption{Generated AFHQ samples obtained by fine-tuning with the unweighted Haar wavelet loss under different EDM formulations.}
  \label{fig:afhq_haar}
\end{figure}

\begin{figure}[t]
  \centering
  \begin{subfigure}{\columnwidth}
    \centering
    \includegraphics[scale = 0.4]{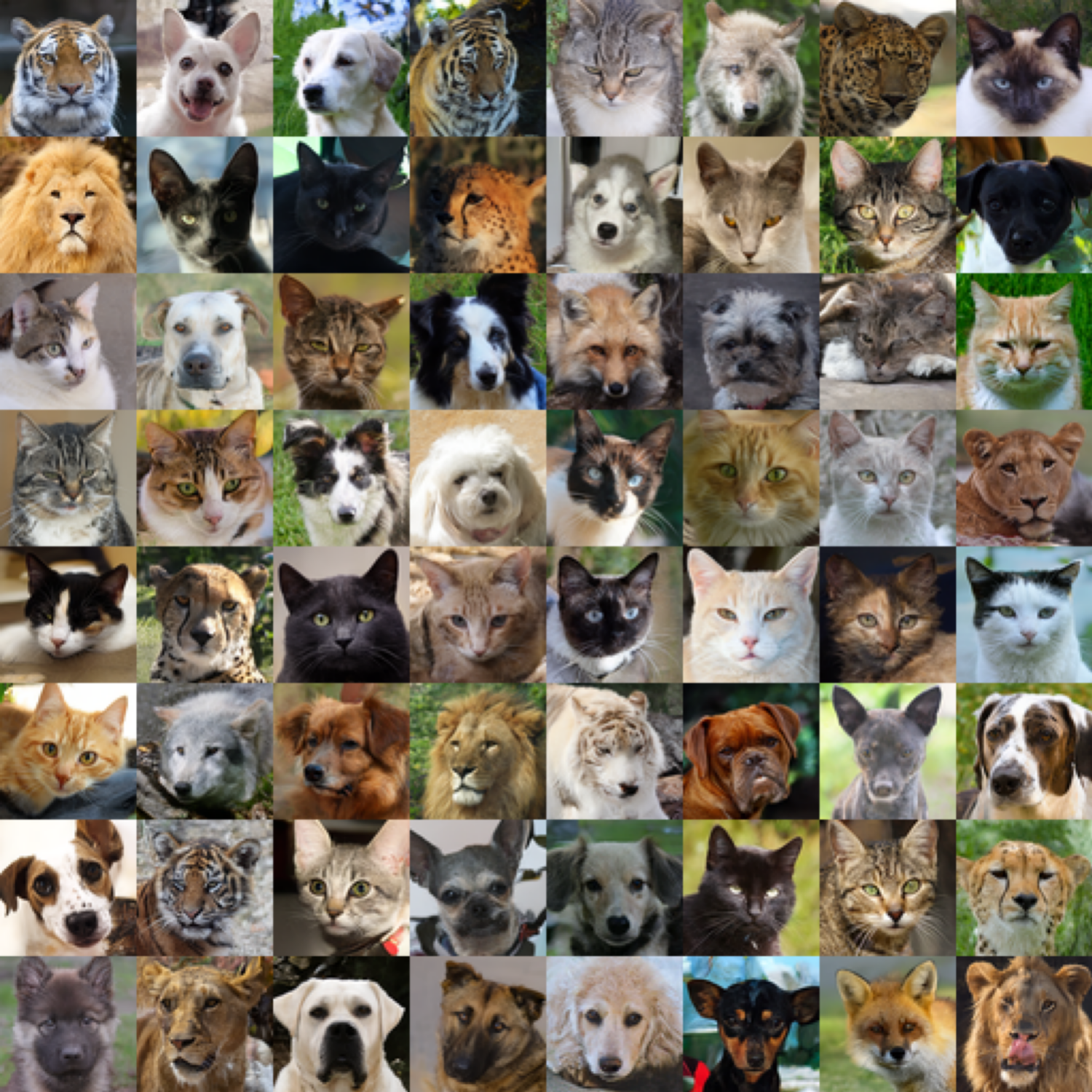}
    \caption{VE-EDM results}
  \end{subfigure}
  \vspace{0.8em}
  \begin{subfigure}{\columnwidth}
    \centering
    \includegraphics[scale = 0.4]{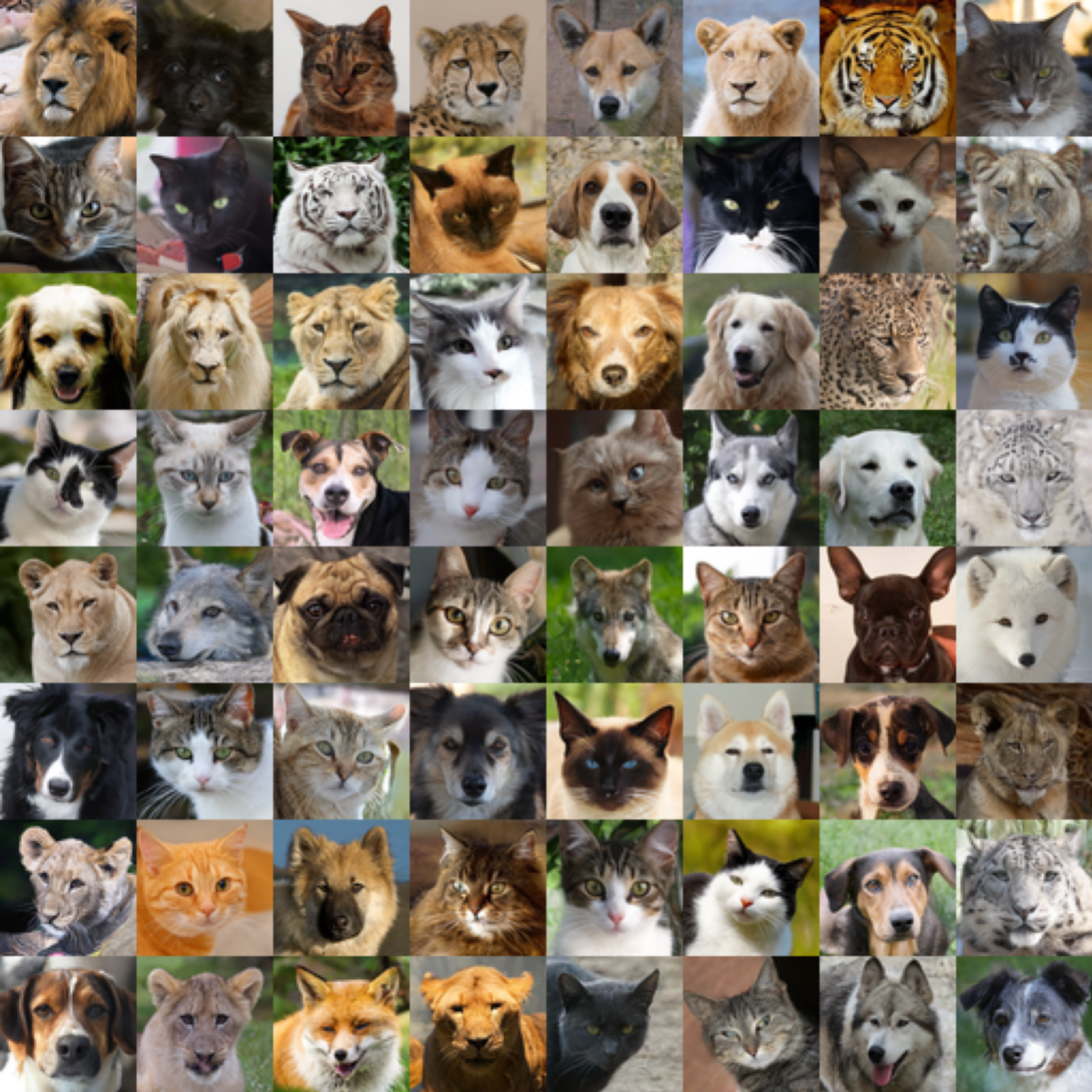}
    \caption{VP-EDM results}
  \end{subfigure}
  \caption{Generated AFHQ samples obtained by fine-tuning with the unweighted bi-orthogonal 1.3 wavelet loss under different EDM formulations.}
  \label{fig:afhq_bior}
\end{figure}

\begin{figure}[t]
  \centering
  \begin{subfigure}{\columnwidth}
    \centering
    \includegraphics[scale = 0.4]{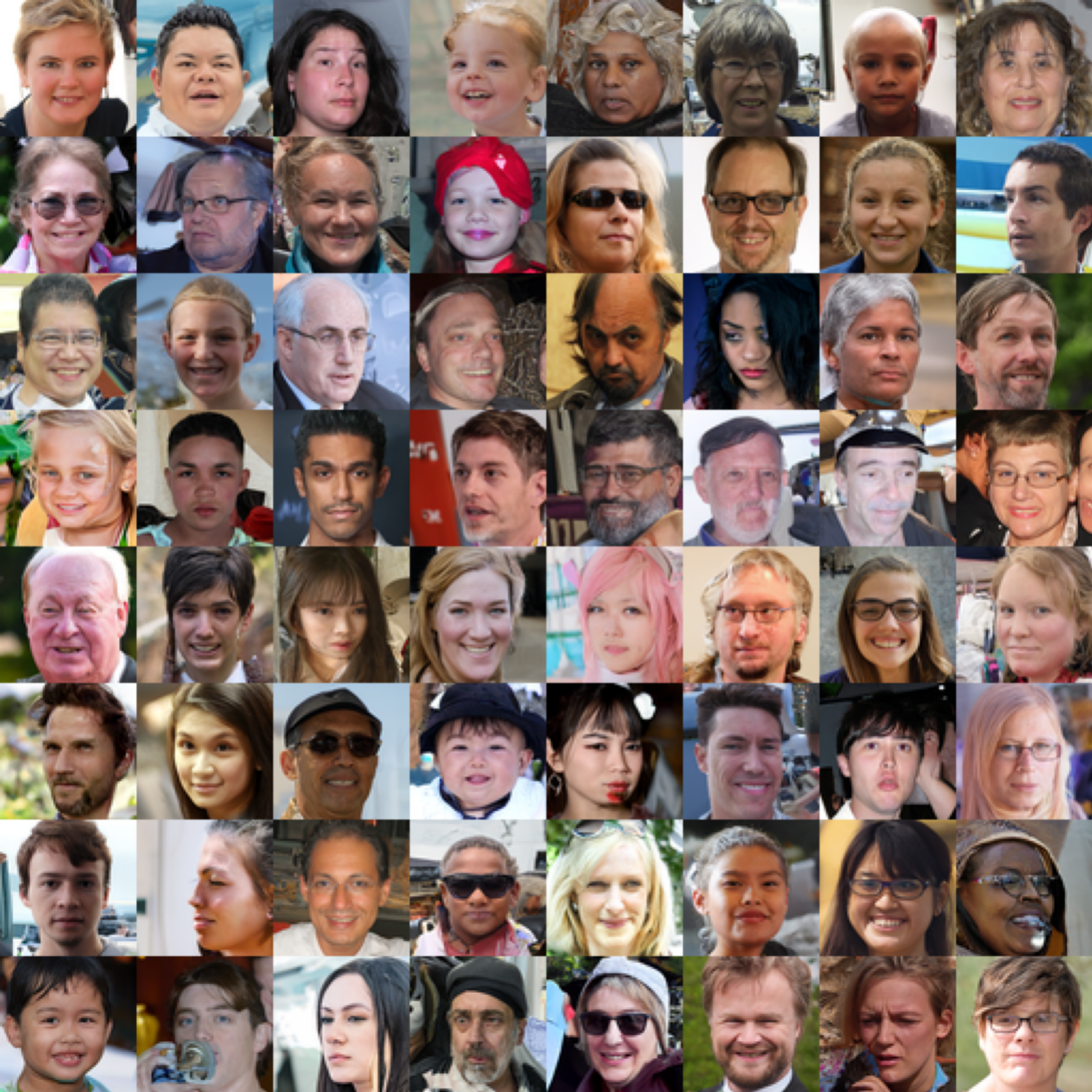}
    \caption{VE-EDM results}
  \end{subfigure}
  \vspace{0.8em}
  \begin{subfigure}{\columnwidth}
    \centering
    \includegraphics[scale = 0.4]{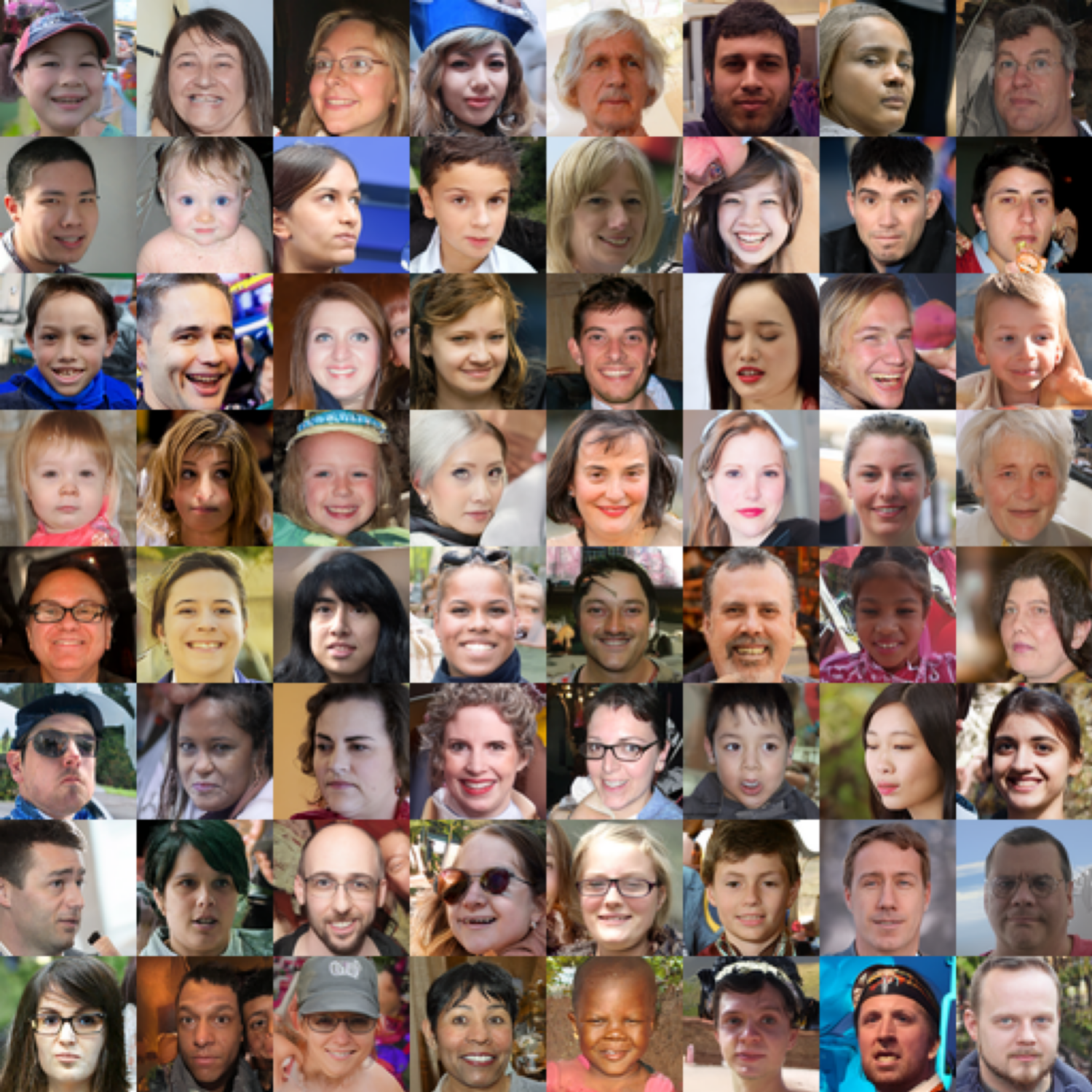}
    \caption{VP-EDM results}
  \end{subfigure}
  \caption{Generated FFHQ samples obtained by fine-tuning with the unweighted Fourier amplitude loss under different EDM formulations.}
  \label{fig:ffhq_amp}
\end{figure}

\begin{figure}[t]
  \centering
  \begin{subfigure}{\columnwidth}
    \centering
    \includegraphics[scale = 0.4]{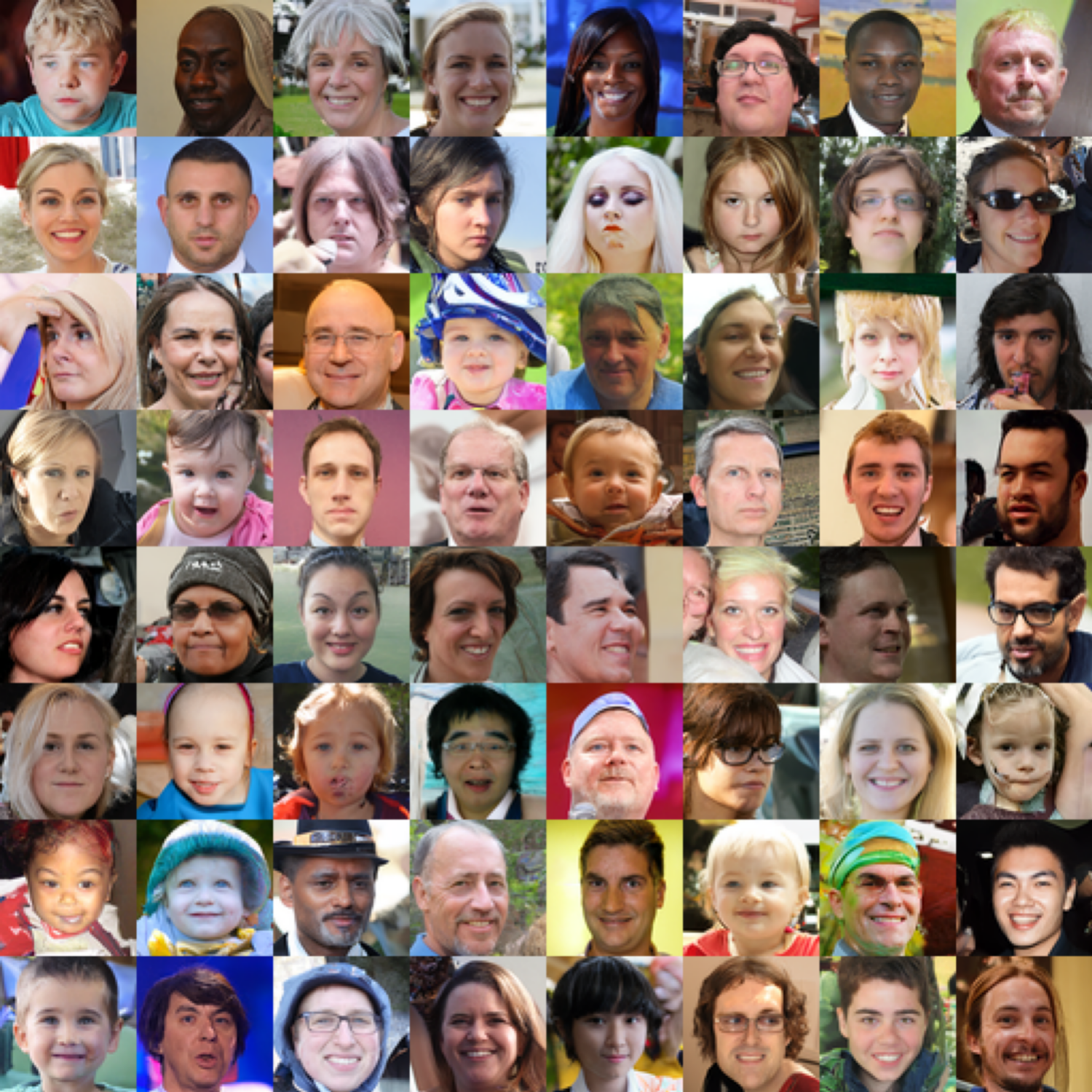}
    \caption{VE-EDM results}
  \end{subfigure}
  \vspace{0.8em}
  \begin{subfigure}{\columnwidth}
    \centering
    \includegraphics[scale = 0.4]{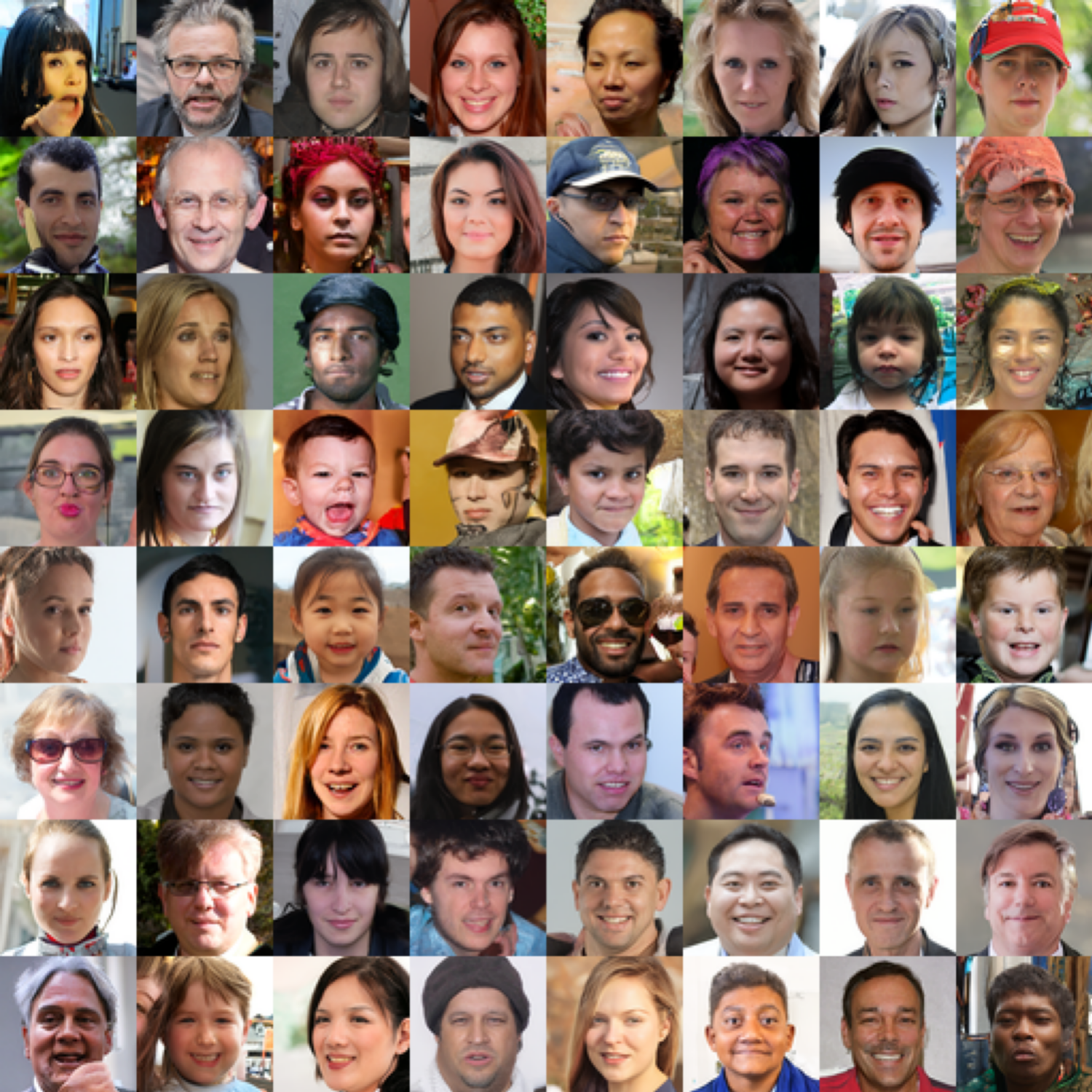}
    \caption{VP-EDM results}
  \end{subfigure}
  \caption{Generated FFHQ samples obtained by fine-tuning with the unweighted Fourier amplitude+phase loss under different EDM formulations.}
  \label{fig:ffhq_amp_phase}
\end{figure}
\begin{figure}[t]
  \centering
  \begin{subfigure}{\columnwidth}
    \centering
    \includegraphics[scale = 0.4]{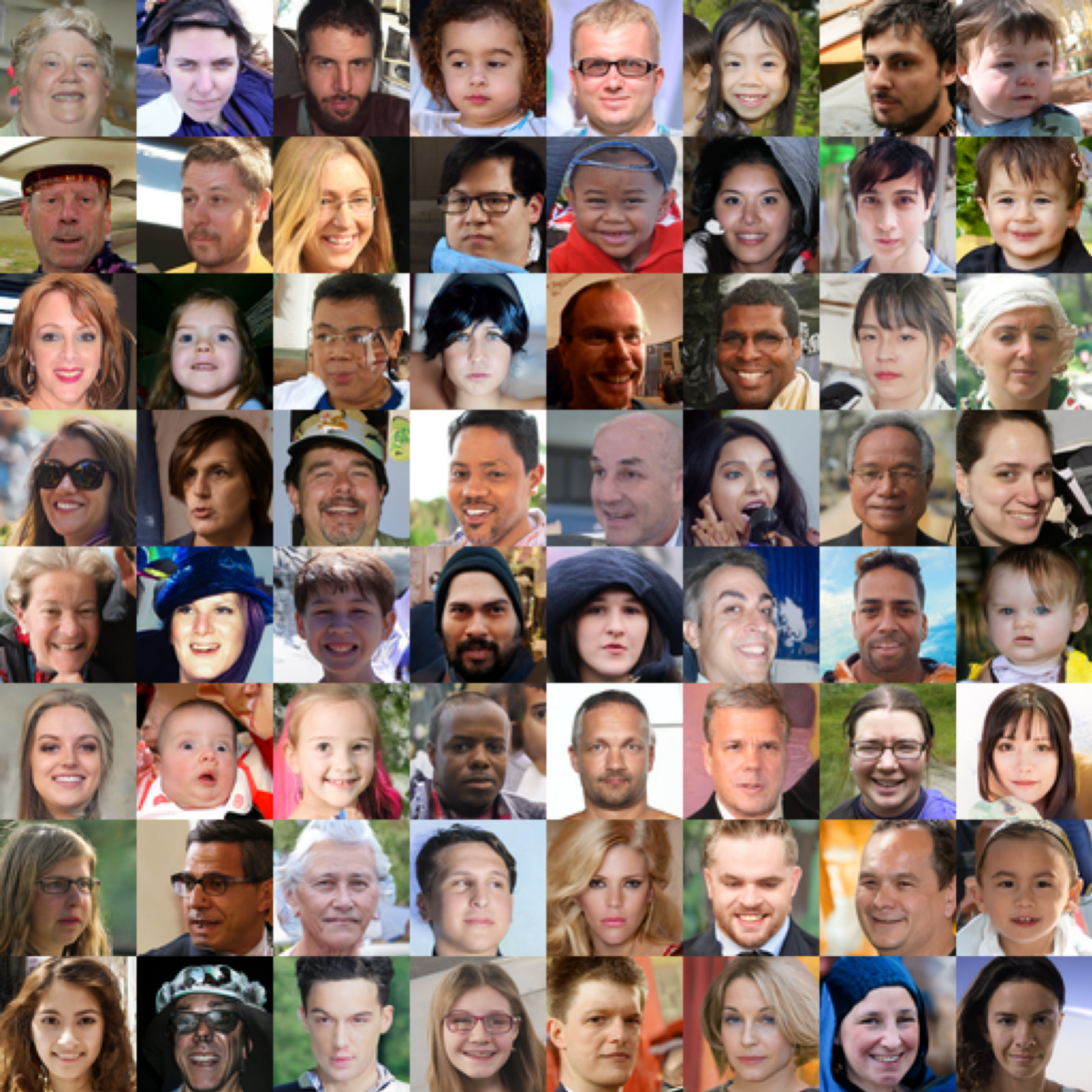}
    \caption{VE-EDM results}
  \end{subfigure}
  \vspace{0.8em}
  \begin{subfigure}{\columnwidth}
    \centering
    \includegraphics[scale = 0.4]{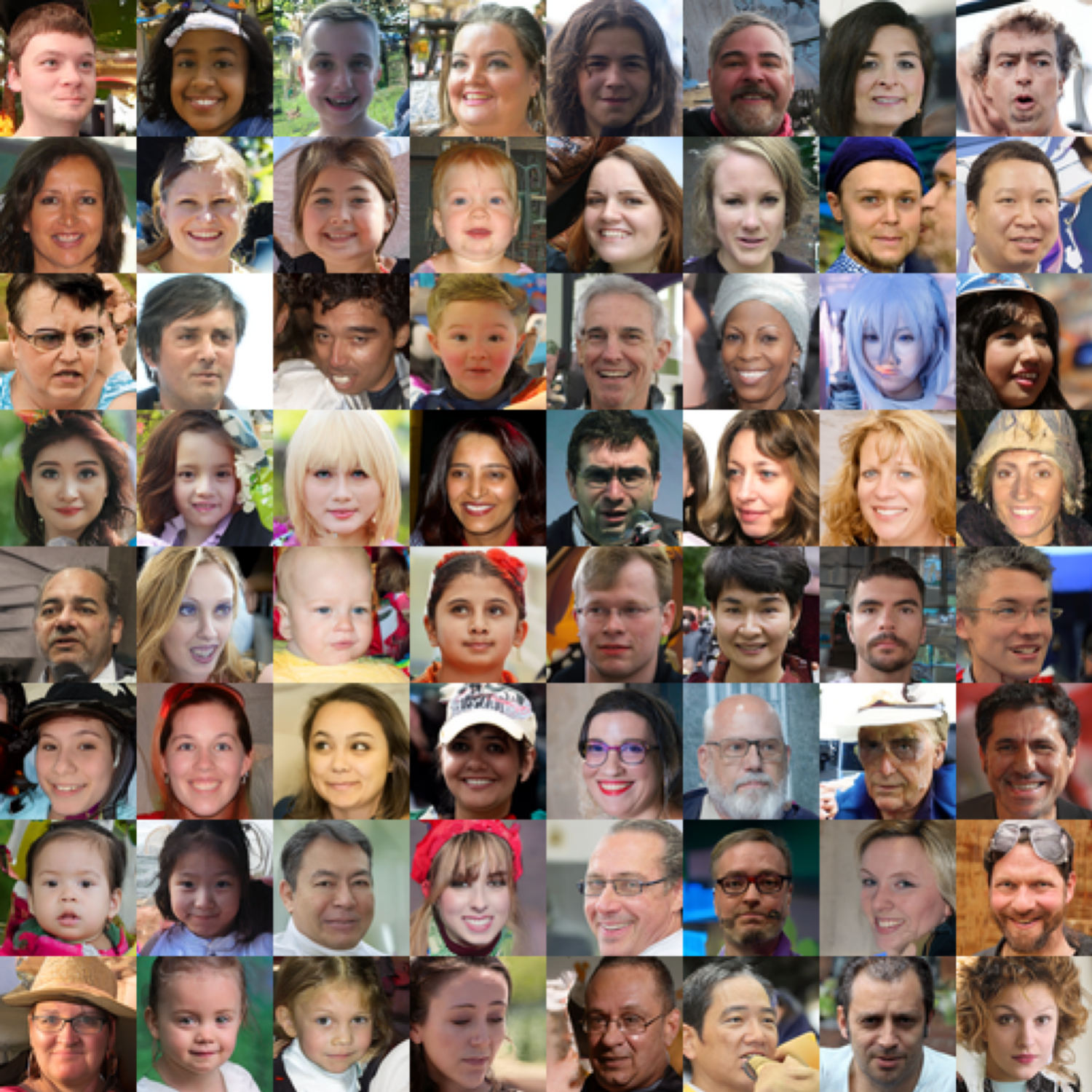}
    \caption{VP-EDM results}
  \end{subfigure}
  \caption{Generated FFHQ samples obtained by fine-tuning with the unweighted Haar wavelet loss under different EDM formulations.}
  \label{fig:ffhq_haar}
\end{figure}

\begin{figure}[t]
  \centering
  \begin{subfigure}{\columnwidth}
    \centering
    \includegraphics[scale = 0.4]{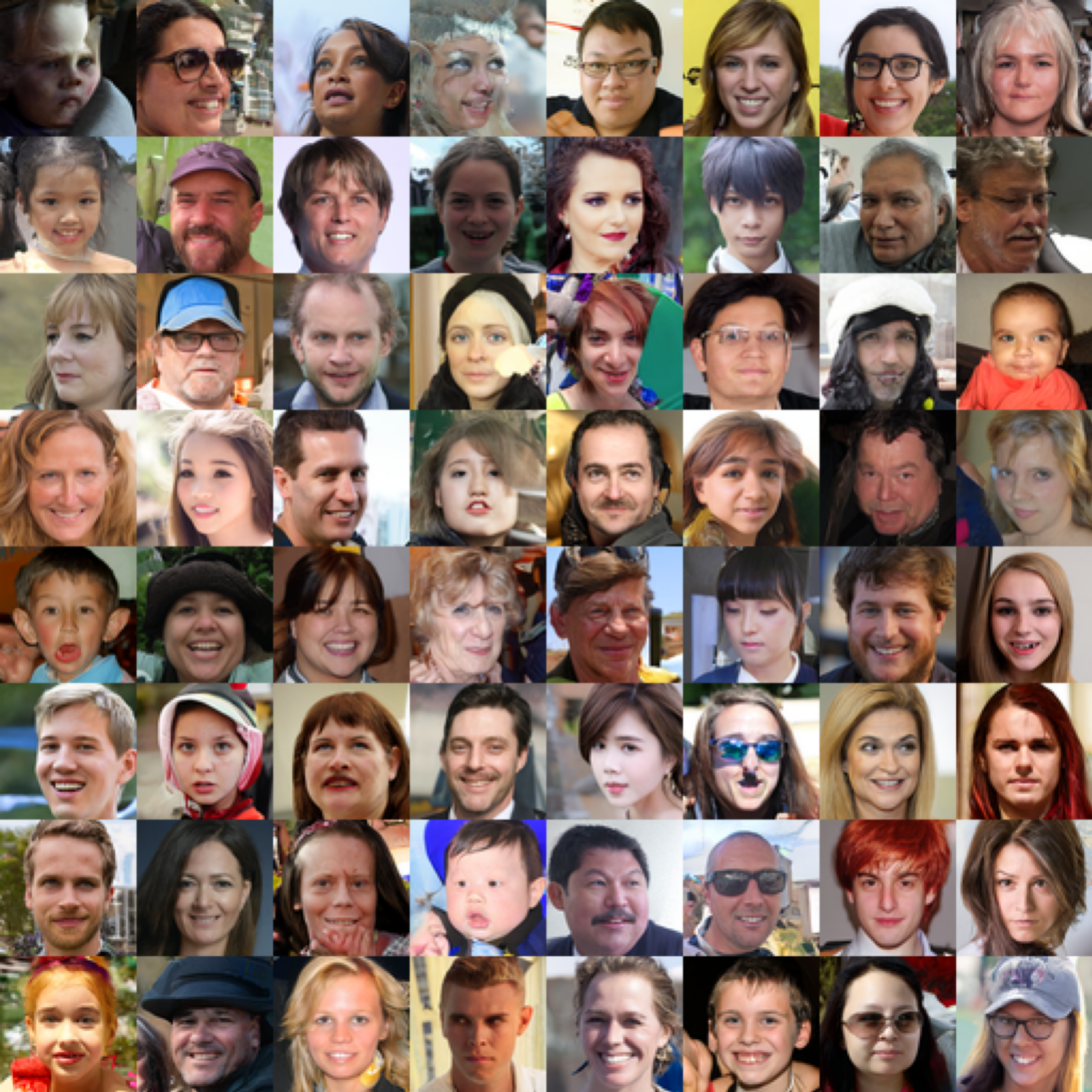}
    \caption{VE-EDM results}
  \end{subfigure}
  \vspace{0.8em}
  \begin{subfigure}{\columnwidth}
    \centering
    \includegraphics[scale = 0.4]{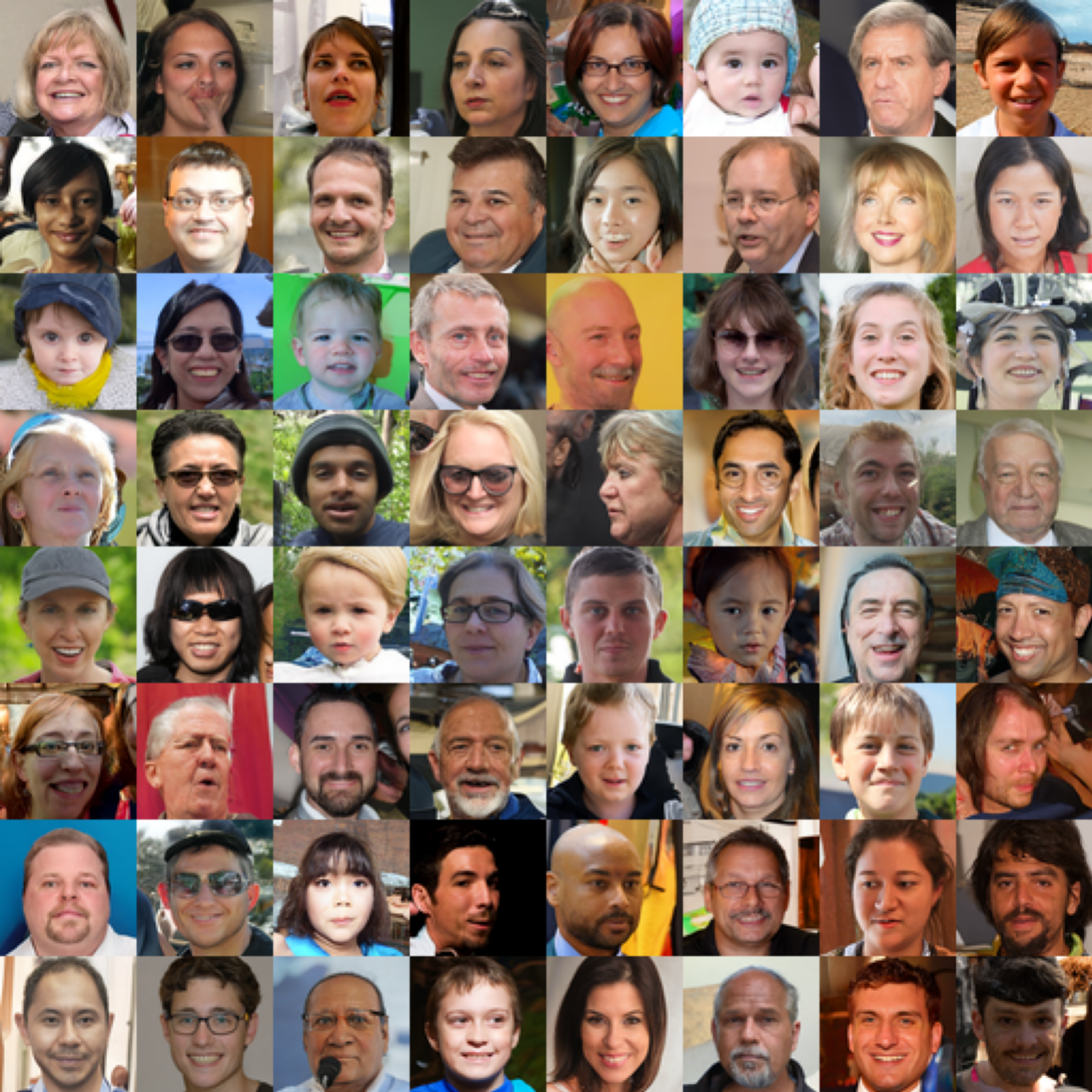}
    \caption{VP-EDM results}
  \end{subfigure}
  \caption{Generated FFHQ samples obtained by fine-tuning with the unweighted bi-orthogonal 1.3 wavelet loss under different EDM formulations.}
  \label{fig:ffhq_bior}
\end{figure}

\section{Generated Images for CIFAR-10 and FFHQ from Distillation} \label{app: gen_image_afhq_ffhq_distill}
Here we present some selected image samples from our models obtained from the Fourier Amplitude $\lambda = 1$ distillation procedure. 
\begin{figure}[t]
  \centering
  \begin{subfigure}{\columnwidth}
    \centering
    \includegraphics[scale = 1.5]{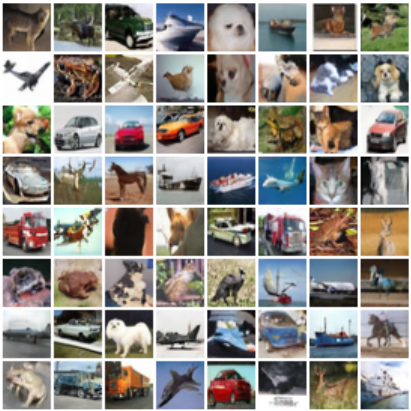}
    \caption{CIFAR-10 Results}
  \end{subfigure}
  \vspace{0.8em}
  \begin{subfigure}{\columnwidth}
    \centering
    \includegraphics[scale = 0.8]{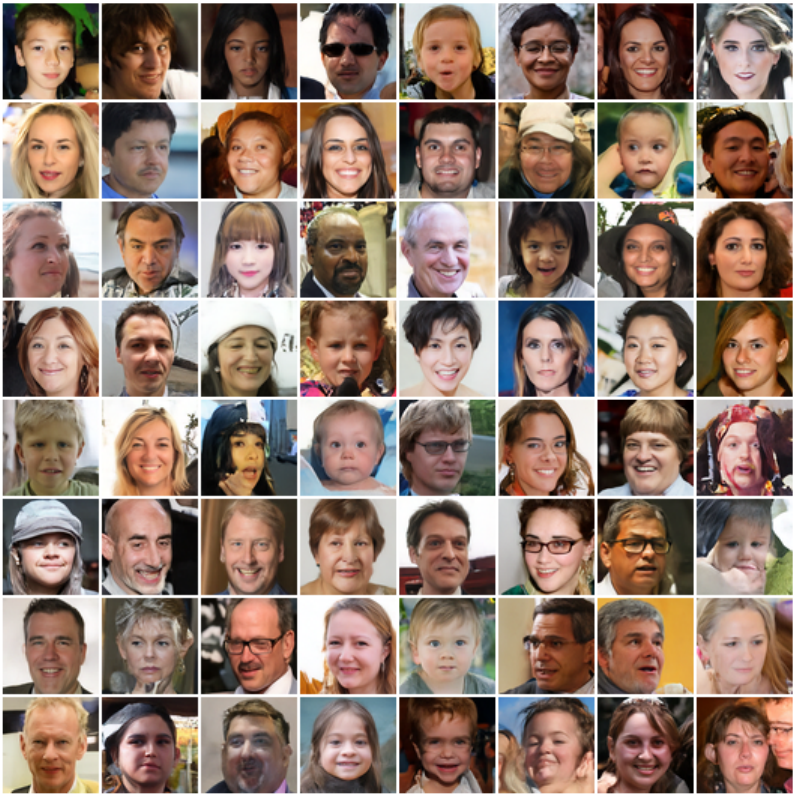}
    \caption{FFHQ Results}
  \end{subfigure}
  \caption{Generated CIFAR-10 and FFHQ samples obtained via SFD distillation with $\lambda = 1$.}
  \label{fig:distill_images}
\end{figure}

\end{document}